\documentclass[11pt]{article}
\usepackage[margin=1in]{geometry}

\usepackage{amsmath,amssymb,amsthm,mathtools}
\usepackage{graphicx}
\usepackage{enumitem}
\usepackage{hyperref}
\hypersetup{hidelinks}
\usepackage{cleveref}
\usepackage{microtype}
\usepackage[numbers]{natbib}

\usepackage{algorithm,algorithmic}
\usepackage{booktabs}
\usepackage{tabularx}
\usepackage{multirow} 
\usepackage{subcaption}
\usepackage{wrapfig}

\usepackage{xspace}

\usepackage{xcolor}

\definecolor{iorange}{RGB}{255, 140, 0}
\definecolor{iblue}{RGB}{30,144,255}
\definecolor{darkgreen}{RGB}{0, 120, 60}
\definecolor{revisionblue}{RGB}{0, 0, 0}
\colorlet{blue}{black}
\colorlet{red}{black}
\colorlet{iorange}{black}
\colorlet{iblue}{black}
\colorlet{darkgreen}{black}
\newcommand{\tcrev}[1]{\textcolor{revisionblue}{#1}}

\usepackage{ulem}
\usepackage{xcolor}

\newtheorem{theorem}{Theorem}[section]
\newtheorem{lemma}[theorem]{Lemma}
\newtheorem{proposition}[theorem]{Proposition}
\newtheorem{definition}[theorem]{Definition}
\newtheorem{remark}[theorem]{Remark}
\newenvironment{wrappedremark}[1]{%
  \par\addvspace{\topsep}\refstepcounter{theorem}%
  \noindent\textbf{Remark \thetheorem}\ (#1).\ \itshape\ignorespaces
}{\par\addvspace{\topsep}}

\newtheorem{counterexample}[theorem]{Counterexample}

\newtheorem{corollary}[theorem]{Corollary}

\newenvironment{namedassumption}[1]{%
  \par\medskip\phantomsection\noindent\textbf{Assumption #1.}\ \itshape
}{%
  \par\medskip
}

\usepackage{placeins}

\newcommand{\dt}{{\Delta t}}

\newcommand{\E}{\mathbb{E}}
\newcommand{\R}{\mathbb{R}}

\newcommand{\norm}[1]{\left\lVert #1 \right\rVert}

\newcommand{\beq}{\begin{equation}}
\newcommand{\eeq}{\end{equation}}

\def\E{\mathbb{E}}

\def\tcb{\textcolor{blue}}
\def\argmax{\text{argmax}}
\def\bl{\color{blue}}

\DeclareMathOperator*{\argmin}{\arg\min}

\title{A Model-Free Continuous-Time RL Algorithm
from Discrete-Time Off-Policy Data with End-to-End Guarantees}
\author{Chaorui Yao\textsuperscript{1}\quad
Yuming Paul Zhang\textsuperscript{2}\quad
Lin Yang\textsuperscript{1}\quad
Yuhua Zhu\textsuperscript{3,*}}
\date{}

\begin{document}
\maketitle
\makeatletter
\begingroup
\renewcommand{\@makefntext}[1]{\noindent#1}
\renewcommand{\thefootnote}{}
\footnotetext[0]{\raggedright\textsuperscript{1}Department of Electrical and Computer Engineering, University of California, Los Angeles.\newline
Emails: \texttt{chaorui@ucla.edu}, \texttt{linyang@ee.ucla.edu}.\par
\textsuperscript{2}Department of Mathematics and Statistics, Auburn University.
Email: \texttt{yzhangpaul@auburn.edu}.\par
\textsuperscript{3}Department of Statistics and Data Science, University of California, Los Angeles.
Email: \texttt{yuhuazhu@ucla.edu}.\par
\textsuperscript{*}Corresponding author.}
\endgroup
\makeatother
\begin{abstract}
\tcb{In this paper, we study continuous-time reinforcement learning, where the
state follows an unknown stochastic differential equation but only a fixed
off-policy batch of discrete-time transitions is available. Standard
discrete-time Q-learning becomes problematic as the sampling interval
shrinks, because differences between action values vanish. Continuous-time
alternatives, such as Deep Advantage Updating and martingale \(q\)-learning,
preserve the action signal but jointly learn a value function and an
action-dependent critic. Building on the PhiBE-Q formulation
\cite{companion_ct_q}, we develop a model-free \tcrev{PhiBE policy iteration (PhiBE-PI)} framework
that uses a single state-action critic while preserving the continuous-time
action-ranking scale. At each iteration, the framework learns the critic
directly from sampled increments and uses it to improve the policy. The
same batch is reused throughout, without explicitly estimating the drift
or diffusion.}

\tcb{We further develop an implementation under linear function approximation
and establish a local end-to-end suboptimality guarantee for the learned
policy. Our analysis quantifies how errors from time discretization,
function approximation, and finite-sample policy evaluation propagate
through policy iteration. It accounts for the dependence of successive
policies on the reused data, so the guarantee does not require a fresh
batch at each iteration. Moreover, we identify a bias--variance trade-off:
smaller sampling intervals improve the approximation to the continuous-time
problem but can increase statistical error at a fixed sample size. Thus,
finer time discretization does not necessarily lead to better policy
performance.}
\end{abstract}

\section{Introduction}

\def\dt{\Delta t}
Many reinforcement-learning problems arise from systems that evolve in
continuous time.  Robotic and physical systems evolve between controller
updates; physiological variables such as glucose levels change continuously
while measurements and interventions arrive intermittently
\cite{Cobelli2009,Battelino2019,Emerson2023}; and portfolio and execution
problems are often modeled through controlled diffusions
\cite{fleming_soner_2006,WangZariphopoulouZhou2020}.  The natural objective in
these settings is a continuous-time stochastic-control objective, but the data
available to a learning algorithm are usually sampled.  We study the common
sampled-data interface in which a simulator or logging system provides a fixed
off-policy batch \(\{(s_i,a_i,r_i,s_i')\}_{i=1}^n\), with
\(s_i'\sim \rho_{\dt}(\cdot\mid s_i,a_i)\) generated by holding the action fixed
over an interval of length \(\dt\).  The resulting mismatch is fundamental: the
control objective is defined in continuous time, whereas the learning algorithm
only observes finite-\(\dt\) one-step transitions.

A first and, from an algorithmic viewpoint, very attractive route is to
turn the sampled data back into a known-model continuous-time control
problem.  One may estimate the drift and diffusion, or more generally the
local generator, from the finite-\(\dt\) transitions and then apply
classical stochastic-control machinery to the resulting HJB equation.  When
the generator is known, policy iteration evaluates a value function by
solving a linear PDE and improves the policy by maximizing the Hamiltonian,
or equivalently the continuous-time action score induced by the value
function
\cite{fleming_soner_2006,Howard1960,PutermanBrumelle1979,
SantosRust2004,KerimkulovSiskaSzpruch2020}.  This route is appealing not
only because it preserves the continuous-time Bellman structure, but also
because the resulting known-model PI has strong algorithmic convergence
theory: it is closely related to Newton--Kantorovich iteration and admits
quadratic, superlinear, or exponential-type convergence rates in various
discrete, diffusion, and HJ/HJB settings
\cite{PutermanBrumelle1979,SantosRust2004,
KerimkulovSiskaSzpruch2020,TangTranZhang2025,GuoTranZhang2025,
TranWangZhang2025}.

\tcb{These guarantees, however, rely on an accurate continuous-time model.
As discussed in \cite{Zhu2024PhiBE}, identifying continuous-time dynamics
from discrete-time observations is generally ill-posed: different
continuous-time dynamics may yield the same discrete-time transitions.
Consequently, errors in the recovered model can propagate through policy
evaluation and improvement, leading to a suboptimal policy despite the fast
convergence of known-model PI.  This motivates learning a finite-\(\dt\)
surrogate of the relevant state-action equation directly from sampled
increments, without requiring explicit recovery of the continuous-time
dynamics.}

A model-free alternative avoids estimating the dynamics and instead learns the
policy directly from data. Its most direct form discretizes everything,
treating each sampling interval as one step of a discrete-time MDP and applying
standard off-policy \(Q\)-learning to the logged one-step transitions
\cite{WatkinsDayan1992,sutton_barto_2018}. The appeal is operational: the
critic is a single state-action object, its Bellman equation is written
directly in terms of the logged transition kernel, and off-policy one-step
samples can be reused across policy-evaluation steps. But this reduction distorts the action-ranking signal in the continuous-time limit.  The first action is applied only over
\([0,\dt]\), so \(Q^{\dt}_\pi(s,a)\) differs from \(V^{\dt}_\pi(s)\) by only
\(O(\dt)\).  Consequently the advantage used for greedy improvement vanishes as
\(\dt\to0\) \cite{TallecBlierOllivier2019,DeAsisSutton2024}.  Discrete-time
\(Q\)-learning fits the data interface, but loses the continuous-time action
scale.

\tcb{Another model-free route seeks a continuous-time analogue of the
\(Q\)-function that preserves the action-ranking signal as \(\dt\to0\).
This idea goes back to Baird's advantage updating \cite{Baird1994},
which rescales the vanishing discrete-time advantage to retain the effect
of an action per unit time.  Tallec et al.\ \cite{TallecBlierOllivier2019}
further developed this approach through Deep Advantage Updating (DAU).
For controlled diffusions, Jia and Zhou \cite{JiaZhou2023} developed the
little \(q\)-function and its martingale characterization. However, DAU and martingale \(q\)-learning jointly learn a value
function and an action-dependent critic, rather than using a standalone
state-action critic as in standard \(Q\)-learning.}
Related continuous-time RL work studies finite-horizon LQ learning,
mean-field-game learning, entropy regularization, and the accuracy of
discretely sampled stochastic policies
\cite{BaseiGuoHuZhang2022,GuoHuXuZhang2023,
SzpruchTreetanthiploetZhang2024,JiaOuyangZhang2025}.  These works address
important continuous-time learning regimes, but they do not provide the
one-batch, model-free, \(Q\)-based policy-iteration mechanism targeted here.

The preceding routes fall short in complementary ways.  The model-based route
has the right continuous-time action ranking and, once the generator is known,
can benefit from fast policy-iteration convergence.  Its bottleneck is
upstream: the generator must be known or accurately recovered from
finite-\(\dt\) transitions, and this recovery error can dominate the subsequent
PI iteration.  Discrete-time \(Q\)-learning has the opposite strength: it gives
a standalone off-policy state-action equation tied directly to one-step
samples, but its advantage signal collapses as \(\dt \to 0\).
\tcb{DAU and martingale \(q\)-learning preserve an \(O(1)\) action signal,
but jointly learn a value function and an action-dependent critic.}
These complementary limitations motivate learning the
relevant state-action equation directly from sampled increments.

This leaves a simple question: can sampled one-step data support policy
iteration for continuous-time control without first recovering the model?  
We seek a \(Q\)-based policy iteration method that keeps the operational
simplicity of discrete-time \(Q\)-learning: for each policy, learn a single
state-action critic from a fixed off-policy batch and improve the policy
greedily.  At the same time, the critic should preserve the continuous-time
action-ranking scale, so that greedy improvement remains meaningful as
\(\dt \to 0\).  The method should learn directly from finite-\(\dt\)
increments, \tcb{without fitting the drift and diffusion or learning a
separate value critic.}
The central question is
whether such a sampled-increment \(Q\)-style object can be propagated through
policy iteration with a local end-to-end guarantee for the original
continuous-time control objective.

\tcrev{Recent methods based on PhiBE, a PDE-based Bellman equation, provide the ingredients for such a construction.}
PhiBE uses finite-step moment information from the sampled kernel
\(\rho_{\dt}\) to build continuous-time PDE-based Bellman equations for policy
evaluation, and Optimal-PhiBE extends this idea to value-level optimal control
\cite{Zhu2024PhiBE,ZhuZhangZhang2025OptimalPhiBE}.  Separately, PhiBE-Q
identifies a nondegenerate continuous-time state-action evaluator whose policy
average recovers the value function and whose greedy action agrees with the
Hamiltonian ranking rule \cite{companion_ct_q}.  Together, these results provide
the two structural ingredients needed here: finite-\(\dt\) moment surrogates
identified from one-step samples, and a continuous-time-scale state-action
ranking.

This paper combines these ingredients in the linear function approximation setting.  We
construct a finite-\(\dt\) \tcb{PhiBE-Q} surrogate by replacing the
infinitesimal generator with one-step moment quantities identified from the
logged transition kernel \(\rho_{\dt}\).  The resulting policy-evaluation
equation is estimated directly from the observed increments
\(\Delta s_i/\dt\) and \(\Delta s_i\Delta s_i^\top/(2\dt)\), without fitting the
drift or diffusion as separate functions.  Combined with greedy improvement,
this yields a one-batch off-policy policy-iteration algorithm: the same batch is
reused at every evaluation step, even though later greedy policies are
data-dependent.

Our contributions are threefold.  First, we formulate a one-batch off-policy
\tcrev{PhiBE policy iteration (PhiBE-PI)} framework and give an implementation with
linear function approximation from one-step samples.  Second, we prove a uniform finite-sample
policy-evaluation bound over the feature-induced greedy policy class.  This
uniformity is essential because the same batch is reused across policy
iteration and later greedy policies depend on that batch.  Third, we establish
a local end-to-end continuous-time suboptimality guarantee for the output policy.
The bound separates surrogate-transfer error, linear function approximation error,
one-batch finite-sample evaluation error, and policy-iteration propagation error. 
The guarantee also exposes the central continuous-time tradeoff.  Decreasing
\(\dt\) improves the population surrogate by making the finite-step moments
closer to the infinitesimal generator, but it worsens the finite-sample radius
because generator-scale increment estimators become more variable.  Thus the
analysis quantifies the same bias--variance tension that makes the sampled-data
continuous-time regime nontrivial.

The rest of the paper is organized as follows.  Section~\ref{sec:setup}
formalizes the learning setting and the small-step action-ranking gap.
Section~\ref{sec:algorithm} introduces the PhiBE-Q surrogate and the
one-batch linear policy-iteration algorithm.  Section~\ref{sec:theory} proves
the local end-to-end finite-sample guarantee.  Section~\ref{sec:experiments} presents
numerical evidence, and Section~\ref{sec:conclusion} discusses limitations and
future directions.

\section{Problem setup and the finite-observation gap}
\label{sec:setup}
\subsection{Continuous-time control with discrete-time off-policy data}
We consider controlled diffusions and aim to learn an optimal feedback policy
\(\pi^\ast:\R^d\to\mathcal A\) that maximizes the discounted continuous-time
return.  For a feedback policy \(\pi\), the continuous-time value function is
defined by
\begin{equation}\label{eq:Vpi}
V^\pi(s)
:=
\E\!\left[
\int_{0}^{\infty} e^{-\beta t}\, r(s_t,\pi(s_t))\,dt
\,\Big|\, s_0=s
\right],
\end{equation}
where the state process follows the It\^{o} dynamics
\[
ds_t=b(s_t,a_t)\,dt+\sigma(s_t,a_t)\,dB_t,
\qquad a_t=\pi(s_t),
\qquad \Sigma(s,a):=\sigma(s,a)\sigma(s,a)^\top.
\]
The optimal continuous-time value function \tcb{and its HJB-optimal feedback
policy are characterized in, e.g., \cite{fleming_soner_2006}, as follows:}
\begin{equation}\label{eq:Vstar}
\begin{aligned}
V^\ast(s)
&:=
\sup_\pi V^\pi(s), \quad 
\tcb{\pi^\ast(s)}
&\tcb{\in\arg\max_{a\in\mathcal A}
\Bigl\{r(s,a)+b(s,a)\cdot\nabla V^\ast(s)
+\tfrac12\Sigma(s,a):\nabla^2V^\ast(s)\Bigr\}.}
\end{aligned}
\end{equation}
In continuous-time RL, the coefficients \((b,\sigma)\) are unknown and only \emph{one-step transitions} off-policy data over a fixed step size \(\Delta t\)
under piecewise-constant actions are available. Concretely, we observe an i.i.d.\ one-step dataset
\begin{equation}\label{eq:def-of-data}
    \mathcal D=\{(s_i,a_i,r_i,s_i')\}_{i=1}^n,\qquad
(s_i,a_i)\sim \mu,\quad r_i=r(s_i,a_i),\quad
s_i'\sim \rho_{\Delta t}(\cdot\mid s_i,a_i),
\end{equation}
where \(r:\mathcal S\times\mathcal A\to\mathbb R\) is the instantaneous
reward-rate function, and \(\rho_{\Delta t}(\cdot\mid s,a)\) is the transition
kernel induced by the SDE when the action is held constant on \([0,\Delta t)\)
(equivalently, the solution of the associated Fokker--Planck equation at time
\(\Delta t\); see e.g.~\cite{ZhuZhangZhang2025OptimalPhiBE}).
Throughout, we assume access only to these one-step samples, rather than
continuous-time trajectories, even though the target control problem is
continuous-time.

\subsection{Why a value function alone is not enough: policy iteration needs action ranking}
A standard route to continuous-time optimal control is policy iteration (PI) when
the dynamics are known
\cite{Howard1960,PutermanBrumelle1979,SantosRust2004,KerimkulovSiskaSzpruch2020}.
Given a policy $\pi_k$, one first evaluates $V^{\pi_k}$ by solving the linear PDE
\begin{equation}\label{eq:ct_eval}
\beta V^{\pi_k}(s)= r(s,\pi_k(s)) + b(s,\pi_k(s))\cdot\nabla V^{\pi_k}(s)
+\tfrac12 \Sigma(s,\pi_k(s)):\nabla^2 V^{\pi_k}(s).
\end{equation}
Crucially, policy improvement requires ranking actions at the same state.  The
known-dynamics PI update uses
\tcb{the instantaneous advantage (equivalently, the HJB residual)}
\cite{JiaZhou2023,TallecBlierOllivier2019},
\begin{equation}\label{eq:ct_adv}
\tcb{q^{\pi_k}(s,a)
=r(s,a)+(\mathcal L^aV^{\pi_k})(s)-\beta V^{\pi_k}(s)},\quad \text{where }(\mathcal L^a f)(s)
    :=b(s,a)\cdot\nabla f(s)
      +\tfrac12\Sigma(s,a):\nabla^2f(s). 
\end{equation}
followed by the greedy update
\(\pi_{k+1}(s)\in\arg\max_{a\in\mathcal A}
\tcb{q^{\pi_k}(s,a)}\).
Therefore, even if $V^{\pi_k}$ were known exactly, policy improvement still
requires an explicit action-ranking object at each state.
\tcb{However, because \(b\) and \(\sigma\) are unknown, the score
\(q^{\pi_k}(s,a)\) cannot be evaluated directly from
\eqref{eq:ct_adv}, even if \(V^{\pi_k}\) were known exactly.}
A state-value function $V^{\pi}$ alone is insufficient for control:
an action-dependent object such as $q^\pi(s,a)$ (or, in discrete time, a state-action value function)
is needed for improvement.

\subsection{Two routes: finite-step equations and continuous-time scores}
\label{subsec:two-routes}

\paragraph{Route A: discrete-time MDP PI/Q-learning degenerates as $\Delta t\to 0$.}

A fixed batch of one-step transitions naturally suggests a discrete-time
Bellman equation.  Treat each simulator step as a discrete-time MDP with
per-step reward \(\tilde r(s,a)=r(s,a)\Delta t\) and discount factor
\(\gamma=e^{-\beta\Delta t}\).  For a fixed policy \(\pi\), define
\begin{align}
V^{\Delta t}_\pi(s)
&=\E\!\left[\sum_{k=0}^\infty \gamma^k \tilde r(s_k,a_k)
\mid s_0=s\right], \quad
Q^{\Delta t}_\pi(s,a)
=\tilde r(s,a)+\gamma\,
\E_{s'\sim\rho_{\Delta t}(\cdot\mid s,a)}
\big[V^{\Delta t}_\pi(s')\big],
\end{align}
which satisfy the Bellman relations
\begin{align}
V^{\Delta t}_\pi(s)
&=\E_{a\sim\pi(\cdot\mid s)}
\big[Q^{\Delta t}_\pi(s,a)\big], \quad
Q^{\Delta t}_\pi(s,a)
=\tilde r(s,a)+\gamma
\E_{s',a'}
\big[Q^{\Delta t}_\pi(s',a')\big].
\label{eq:dt_q_bellman}
\end{align}
\tcb{Equation~\eqref{eq:dt_q_bellman} is closed in
\(Q_\pi^{\Delta t}\) and conditions on the same one-step kernel
\(\rho_{\Delta t}(\cdot\mid s,a)\) that generates the data.
Consequently, \(Q_\pi^{\Delta t}\) can be estimated directly from
off-policy one-step transitions, without separately estimating
\(V_\pi^{\Delta t}\), and the fitted \(Q\)-function can be used directly
for greedy policy improvement \cite{WatkinsDayan1992}.}

In discrete time, $Q^{\Delta t}_\pi$ and $V^{\Delta t}_\pi$
are genuinely distinct objects:
$Q^{\Delta t}_\pi(s,a)$ captures the effect of committing
to action $a$ over the first interval $[0,\Delta t)$
before reverting to $\pi$.
However, as $\Delta t\to 0$, this distinction collapses.
The difference between $Q^{\Delta t}_\pi(s,a)$ and $V^{\Delta t}_\pi(s)$
comes only from reward accumulation and state evolution over a time window of length $\Delta t$.
Under standard regularity (bounded rewards and smooth value functions),
both contributions vanish at rate $O(\Delta t)$, yielding
\[
Q^{\Delta t}_\pi(s,a)=V^{\Delta t}_\pi(s)+O(\Delta t).
\]
Consequently, the discrete-time action advantage satisfies
\[
A^{\Delta t}_\pi(s,a)
:=
Q^{\Delta t}_\pi(s,a)-V^{\Delta t}_\pi(s)
=
O(\Delta t),
\]
so greedy improvement becomes increasingly ill-conditioned
as $\Delta t\to 0$
\tcb{\cite{Baird1994,TallecBlierOllivier2019,JiaZhou2023}}.
\tcb{Thus, Route~A is naturally compatible with one-step off-policy data,
but its action-ranking signal becomes too small when the action changes more frequently as \(\Delta t\to0\).}

\paragraph{Route B: continuous-time \(q\)-learning has the right scale, but leaves open practical \(Q\)-like algorithm design.}
Continuous-time \(q\)-learning takes the complementary route.  It starts from
the observation that the discrete-time advantage is only \(O(\Delta t)\), and
rescales this vanishing advantage to keep its first-order term.  \tcb{This
idea originates from advantage updating proposed by Baird \cite{Baird1994}
and was further developed by Tallec et al.\ \cite{TallecBlierOllivier2019}
for deterministic system through Deep Advantage Updating (DAU).  Jia and Zhou \cite{JiaZhou2023}
developed the little \(q\)-function and its martingale characterization for general
controlled diffusions.  In the stationary, unregularized setting considered
here, this function may be viewed as}
\[
    q^\pi(s,a)
    =
    \lim_{\Delta t\to0}
    \frac{Q^{\Delta t}_\pi(s,a)-V^{\Delta t}_\pi(s)}{\Delta t}.
\]
\tcb{Equivalently, when \(V^\pi\) is smooth, \(q^\pi\) satisfies
\eqref{eq:ct_adv}, with \(\pi_k\) replaced by \(\pi\).}
\tcb{Thus,} \(q^\pi\) preserves the \(O(1)\) action sensitivity that
\(Q^{\Delta t}_\pi\) loses as \(\Delta t\to0\)
\cite{JiaZhou2023,TallecBlierOllivier2019}.

\tcb{The algorithmic limitation is twofold.  First, these methods require
coupled learning of a value function and an action-dependent critic.  DAU
jointly learns \(V_\theta\) and a rescaled advantage \(A_\psi\) through the
parameterization
\[
    Q_\Theta(s,a)=V_\theta(s)+\Delta t\,A_\psi(s,a),
\]
using off-policy one-step transitions.  Jia and Zhou \cite{JiaZhou2023}
jointly parameterize \(V_\theta\) and \(q_\psi\) and learn both functions
from behavior-policy trajectories through a martingale characterization,
subject to a policy-dependent compatibility condition.  Neither
construction provides a closed one-step evaluation equation for the
action-dependent critic alone.}

\tcb{Second, the direct off-policy optimal-\(q\) construction of
Jia and Zhou \cite{JiaZhou2023} does not proceed through iterative policy
improvement.  Instead, it jointly characterizes \(V^\ast\) and
\(q^\ast\) while requiring the \(q\)-approximator to satisfy
\[
Z_\psi(s)
:=
\int_{\mathcal A}
\exp\{q_\psi(s,a)/\lambda_{\rm ent}\}\,da
=1
\qquad\text{for every }s\text{ and }\psi.
\]
Constructing a generic function class that satisfies this statewise
normalization is difficult.}

Together, the two routes leave a concrete gap.  Route~A matches the one-step off-policy data, but its action signal vanishes as \(\Delta t\to0\).  Route~B keeps the continuous-time
action signal, but does not give an autonomous one-step \(Q\)-equation tied to
the logged kernel \(\rho_{\Delta t}\).  We need a continuous-time Bellman object
that has both features: the right action scale and the standard role of a
state-action critic learned from one-step off-policy data.

\subsection{\tcb{Continuous-time \(Q\)-functions and their PhiBE approximation}}
\label{subsec:desired-object}
\label{sec:ctQ}
\label{sec:phibeq}

\tcb{The discussion above points to a short wish list.  We seek a
\(Q\)-function that 1) keeps the action signal \(O(1)\), 2) avoids a separate
value critic, and 3) can be learned from the one-step off-policy data.  The PhiBE-Q framework \cite{companion_ct_q} provides
a starting point: its continuous-time state-action function retains the
action ranking of \(q^\pi\), while its policy average recovers \(V^\pi\).
This identity allows the evaluation equation to be written in terms of
\(Q^\pi\) alone.}

\tcb{For any state-action function \(Q\), write its policy average as
\(\bar Q^\pi(s):=\mathbb E_{a\sim\pi(\cdot\mid s)}[Q(s,a)]\).
State derivatives act on the entire policy average, including the state
dependence of \(\pi\).}
{\color{blue}
\begin{definition}[Continuous-time \(Q\)-function]
\label{def:ctQ}
For an admissible policy \(\pi\) with smooth discounted value \(V^\pi\),
define
\begin{equation}\label{eq:ctQ}
    \beta Q_{\mathrm{ct}}^\pi
    =r+\mathcal L^a\bar Q_{\mathrm{ct}}^\pi, \quad \bar Q^\pi(s):=\mathbb E_{a\sim\pi(\cdot\mid s)}[Q(s,a)]
\end{equation}
\end{definition}
}

\tcb{Note that under well-posedness of the policy evaluation equation \eqref{eq:ct_eval}, by taking expectation of the above equation over $a\sim \pi$, one has the policy average identity \(\bar Q_{\mathrm{ct}}^\pi=V^\pi\).  Thus, policy evaluation
requires only a single \(Q\)-critic, with the value function recovered by
policy averaging.  Moreover, \eqref{eq:ct_adv} gives
\(Q_{\mathrm{ct}}^\pi=V^\pi+\beta^{-1}q^\pi\), so it preserves the action
ranking of \(q^\pi\), with an action-dependent component independent of
\(\Delta t\).}

\tcb{However, in our setting the dynamics are unknown, and only
discrete-time off-policy transition data are available.  We therefore use
the PhiBE construction to replace the unknown generator coefficients by
conditional moments of the observed one-step increments
\cite{Zhu2024PhiBE,ZhuZhangZhang2025OptimalPhiBE}.}
{\color{blue}
\begin{definition}[PhiBE-\(Q\) approximation]
\label{def:phibeQ}
For the transition kernel \(\rho_{\Delta t}\), define
\begin{equation}\label{eq:moment_surrogates}
\begin{aligned}
    b_{\Delta t}(s,a)
    &:=\mathbb E_{s'\sim\rho_{\Delta t}(\cdot\mid s,a)}
        \left[\frac{s'-s}{\Delta t}\right],\quad
    \Sigma_{\Delta t}(s,a)
    &:=\mathbb E_{s'\sim\rho_{\Delta t}(\cdot\mid s,a)}
        \left[\frac{(s'-s)(s'-s)^\top}{\Delta t}\right].
\end{aligned}
\end{equation}
Let \(\mathcal L_{\Delta t}^a\) denote the operator obtained from
\(\mathcal L^a\) by replacing \(b\) and \(\Sigma\) with
\(b_{\Delta t}\) and \(\Sigma_{\Delta t}\).  For an admissible policy
\(\pi\), write \(r^\pi(s):=\mathbb E_{a\sim\pi(\cdot\mid s)}[r(s,a)]\)
and \(\mathcal L_{\Delta t}^\pi f
:=\mathbb E_{a\sim\pi(\cdot\mid s)}[\mathcal L_{\Delta t}^a f]\).
Under the well-posedness assumptions stated below, let
\(S_t^{\Delta t,\pi}\) be the surrogate process with generator
\(\mathcal L_{\Delta t}^\pi\), and define its discounted value by
\begin{equation}\label{eq:surrogate_value_return}
    V^\pi_{\Delta t}(s)
    :=\mathbb E_s\!\left[
        \int_0^\infty e^{-\beta t}
            r^\pi(S_t^{\Delta t,\pi})\,dt
    \right].
\end{equation}
The PhiBE-\(Q\) approximation for policy \(\pi\) is the solution of
\begin{equation}\label{eq:phibeQ}
    \beta Q_{\Delta t}^\pi
    =r+\mathcal L_{\Delta t}^a\bar Q_{\Delta t}^\pi.
\end{equation}
\end{definition}
}

\section{\tcrev{PhiBE-PI}}
\label{sec:algorithm}

We develop an abstract \tcrev{PhiBE-PI} framework based on the
PhiBE-Q evaluator, using a single off-policy batch for repeated
policy evaluation and improvement.  The framework accommodates different
function approximations and policy updates.  We propose a concrete
instantiation combining linear Galerkin evaluation with hard-greedy
improvement and rigorously analyze it in Section~\ref{sec:theory}.

\subsection{An Abstract Policy-Iteration Framework}
\label{subsec:abstract_pi_template}

There are different ways to solve the PhiBE-Q equation
\eqref{eq:phibeQ}.  Rather than commit to a specific function class and
evaluation procedure up front, we describe a policy-iteration template
that decouples the algorithmic structure from the choice of approximator.

The framework alternates between two oracles:
\(\textsc{Eval}(\mathcal D,\pi)\) fits \(Q^\pi_{\Delta t}\) from the
batch through \eqref{eq:phibeQ}, and
\(\textsc{Improve}(\pi,\widehat Q^\pi_{\Delta t})\) updates the policy
using the fitted evaluator.

\begin{algorithm}[H]
\caption{Abstract \tcrev{PhiBE-PI}}
\label{alg:abstract_pi}
\begin{algorithmic}[1]
\REQUIRE Fixed batch \(\mathcal D\), initial policy \(\pi_0\), oracles
\(\textsc{Eval}\) and \(\textsc{Improve}\), iteration budget \(N\).
\FOR{\(k=0,\ldots,N-1\)}
    \STATE \(\widehat Q^{\pi_k}_{\Delta t}
        \leftarrow\textsc{Eval}(\mathcal D,\pi_k)\).
    \STATE \(\pi_{k+1}\leftarrow
        \textsc{Improve}(\pi_k,\widehat Q^{\pi_k}_{\Delta t})\).
\ENDFOR
\RETURN \(\pi_N\).
\end{algorithmic}
\end{algorithm}

\paragraph{$\textsc{Eval}(\mathcal D,\pi_k)$: Evaluation from sampled residuals .}
For \((s_i,a_i,r_i,s_i')\in\mathcal D\), let
\(\Delta s_i=s_i'-s_i\).  The sampled residual of \eqref{eq:phibeQ} is
\begin{equation}\label{eq:sample-residual}
    \delta_i^\pi(\tcb{Q_{\dt}})
    :=\beta \tcb{Q_{\dt}}(s_i,a_i)-r_i
      -\frac{\Delta s_i}{\Delta t}\cdot\nabla_s\tcb{\bar Q^\pi_{\dt}}(s_i)
      -\frac{\Delta s_i\Delta s_i^\top}{2\Delta t}
          :\nabla_s^2\tcb{\bar Q^\pi_{\dt}}(s_i).
\end{equation}
For fixed \(\pi\) and \(\tcb{Q_{\dt}}\), its conditional expectation given \((s_i,a_i)\)
equals the residual of \eqref{eq:phibeQ}.  This permits direct use of the
logged increments without separately estimating the moment coefficients.
\tcb{One implementation of \(\textsc{Eval}(\mathcal D,\pi)\) could be to choose a function class \(\mathcal F\) and testing
features \(\psi_{Q_{\dt}}\), and then approximately solve}
\begin{equation}\label{eq:projection-residual}
    \frac1n\sum_{i=1}^n
        \tcb{\psi_{Q_{\dt}}}(s_i,a_i)\delta_i^\pi(\tcb{Q_{\dt}})=0,
    \qquad \tcb{Q_{\dt}}\in\mathcal F.
\end{equation}

\tcb{We make several remarks} regarding the above vanilla version of $\textsc{Eval}(\mathcal D,\pi_k)$.
Regarding the calculation of the sampled residual \eqref{eq:sample-residual}, first, when policy averages and their derivatives cannot be evaluated analytically,
they can be estimated from target-policy action samples as described in
Appendix~\ref{app:policy-average-derivatives}.
\tcb{Second, when analytic state derivatives are unavailable, the
policy average can be evaluated at neighboring states and differentiated
numerically using finite differences; see \cite{ChoKim2026} for a
related policy-iteration scheme.} This also provides a numerical
update when classical derivatives do not exist, but the guarantees below
require the stated smoothness assumptions and do not cover that case
or the additional finite-difference error. \tcb{Third, regarding the implementation of \(\textsc{Eval}\), the interface also admits other fitting procedures and nonlinear function approximations.
For example, an iterative method similar to TD takes the form
\(\theta_{j+1}=\theta_j-\alpha_j\delta_i^\pi(Q_{\dt,\theta_j})\nabla_\theta Q_{\dt,\theta_j}(s_i,a_i)\).
Alternatively, one may minimize the empirical squared residual:
\(\min_\theta\frac1n\sum_{i=1}^n[\delta_i^\pi(Q_{\dt,\theta})]^2\).}

\paragraph{$\textsc{Improve}(\pi_k,\widehat Q^{\pi_k}_{\Delta t})$: Policy improvement.}
\tcb{The improvement oracle may maximize the fitted \(Q\)-function through the greedy update
\(\pi_{k+1}(s)\in\arg\max_{a\in\mathcal A}\widehat Q^{\pi_k}_{\Delta t}(s,a)\),
use a softmax update
\(\pi_{k+1}(a\mid s)=\exp\{\widehat Q^{\pi_k}_{\Delta t}(s,a)/\tau\}/\int_{\mathcal A}\exp\{\widehat Q^{\pi_k}_{\Delta t}(s,a')/\tau\}\,da'\)
with temperature \(\tau>0\), or incorporate the current policy through a conservative update
\(\pi_{k+1}(\cdot\mid s)=(1-\eta)\pi_k(\cdot\mid s)+\eta\widetilde\pi_{k+1}(\cdot\mid s)\),
where \(\eta\in(0,1]\) and \(\widetilde\pi_{k+1}\) is a candidate policy obtained by greedy or softmax improvement.}
Algorithm~\ref{alg:abstract_pi} leaves both oracle choices open and reuses
the same batch throughout the iteration.

\subsection{Linear Features and Hard-Greedy Improvement}
\label{subsec:linear_phibeq_pi}

We instantiate the framework with linear residual projection for policy
evaluation and hard-greedy improvement.

\paragraph{Linear Galerkin evaluation.}
\label{par:linear-galerkin-policy-evaluation}
\tcb{For fixed features \(\Phi:\mathbb R^d\times\mathcal A\to\mathbb R^{d_\phi}\),
set \(Q_{\dt,\theta}(s,a)=\Phi(s,a)^\top\theta\) and write
\(\Phi^\pi(s):=\mathbb E_{\pi(\cdot\mid s)}\Phi(s,a)\), so that
\(\bar Q_{\dt,\theta}^\pi(s)=\Phi^\pi(s)^\top\theta\).
Taking \(\psi_{Q_{\dt}}=\Phi\) in \eqref{eq:projection-residual}
gives the Galerkin system \(\widehat A^\pi_{\Delta t}\theta=\widehat c\), where}
\begin{align}
    \widehat A^\pi_{\Delta t}
    &:=\frac1n\sum_{i=1}^n\Phi(s_i,a_i)
      \left[
        \beta\Phi(s_i,a_i)
        -\frac{\Delta s_i}{\Delta t}\cdot\nabla_s\Phi^\pi(s_i)
        -\frac{\Delta s_i\Delta s_i^\top}{2\Delta t}
            :\nabla_s^2\Phi^\pi(s_i)
      \right]^\top,
        \label{eq:Ahat}\\
    \widehat c
    &:=\frac1n\sum_{i=1}^n r_i\Phi(s_i,a_i).
        \label{eq:chat}
\end{align}
Only \(\widehat A^\pi_{\Delta t}\) depends on the policy;
\(\widehat c\) is computed once.
\tcb{When the matrix is nonsingular, the solution is
\(\widehat\theta^\pi_{\Delta t}=(\widehat A^\pi_{\Delta t})^{-1}\widehat c\),
with fitted evaluator \(\widehat Q^\pi_{\Delta t,\Phi}=\Phi^\top\widehat\theta^\pi_{\Delta t}\).
As in LSTD~\cite{BradtkeBarto1996}, the fitted residual is empirically orthogonal to the features,
with the sample PhiBE-Q residual replacing the standard Bellman residual.}
A fixed fallback, such as retaining the current policy, may be used for
a singular system.

\paragraph{Hard-greedy iteration.}
\tcb{Since positive rescaling preserves the maximizing actions, policies can
be indexed by coefficient directions \(u\in\mathbb S_\Phi:=S^{d_\phi-1}\).
For each \(u\), fix a measurable selector
\(\pi_u(s)\in\arg\max_a\Phi(s,a)^\top u\).
Algorithm~\ref{alg:phibeq_pi} uses normalized fitted coefficients to update the policy.}

\begin{algorithm}[H]
\caption{\tcrev{PhiBE-PI:} Linear Galerkin evaluation with hard-greedy improvement}
\label{alg:phibeq_pi}
\begin{algorithmic}[1]
\REQUIRE Fixed batch \(\mathcal D\), step size \(\Delta t\), discount
\(\beta\), features \(\Phi\), initial direction \(u_0\in\mathbb S_\Phi\),
iteration budget \(N\).
\STATE Set \(\pi_0=\pi_{u_0}\) and form \(\widehat c\) using \eqref{eq:chat}.
\FOR{\(k=0,\ldots,N-1\)}
    \STATE \tcb{Assemble \(\widehat A^{\pi_k}_{\Delta t}\) using \eqref{eq:Ahat} and \(\mathcal D\).}
    \STATE Solve
    \(\widehat A^{\pi_k}_{\Delta t}\widehat\theta^{\pi_k}_{\Delta t}
        =\widehat c\).
    \STATE Set
    \[
        u_{k+1}:=
        \begin{cases}
            \widehat\theta^{\pi_k}_{\Delta t}/
            \|\widehat\theta^{\pi_k}_{\Delta t}\|_2,
                &\widehat\theta^{\pi_k}_{\Delta t}\ne0,\\
            u_k,&\text{otherwise},
        \end{cases}
        \qquad \pi_{k+1}:=\pi_{u_{k+1}}.
    \]
\ENDFOR
\RETURN \(\pi_N\).
\end{algorithmic}
\end{algorithm}

\section{Local End-to-End Error Decomposition and Main Guarantee}
\label{sec:theory}

We analyze the linear-feature, hard-greedy algorithm in
Algorithm~\ref{alg:phibeq_pi}.  We first bound its policy-evaluation error,
separating the approximation error from using linear features from the
finite-sample error in estimating the Galerkin system.  We then show how
these errors propagate through greedy policy updates.  Finally, we
combine this policy-iteration bound with the error from replacing the
continuous-time dynamics by the Optimal-PhiBE surrogate to obtain an
end-to-end control guarantee.

We use the continuous-time values \(V^\pi,V^*\) from
\eqref{eq:Vpi} and \eqref{eq:Vstar}, and the surrogate policy value
\(V^\pi_{\Delta t}\) defined in \eqref{eq:surrogate_value_return}.
The surrogate optimal value \(V^*_{\Delta t}\) is the supremum of
discounted returns over admissible controls in the same surrogate problem.
Its HJB characterization is the one in
\cite[Sections~2.1 and 3.2]{ZhuZhangZhang2025OptimalPhiBE}, applied to
the surrogate coefficients:
\begin{equation}\label{eq:optimal_phibe_hjb}
    \beta V^*_{\Delta t}(s)
    =\sup_{a\in\mathcal A}
       \bigl\{r(s,a)+(\mathcal L_{\Delta t}^aV^*_{\Delta t})(s)\bigr\}.
\end{equation}

Lemma~\ref{lem:pi-optimal-value-regularity} verifies classical
regularity under the standing assumptions below.

\subsection{Master decomposition and stochastic transfer}
\label{subsec:master-decomp}
\label{subsec:stoch-transfer}

{\color{blue}
\paragraph{Local performance criterion.}
We evaluate the learned policy from initial states in a ball $B_R$,
for a fixed $R>0$. Let $\varrho$ be any bounded probability density with
\[
\operatorname{supp}(\varrho)\subset B_R,
\qquad
\|f\|_\varrho^2:=\int_{\mathbb R^d}|f(s)|^2\varrho(s)\,ds.
\label{eq:target-norm-compatibility}\tag{TNC}
\]
Thus $\|V^*-V^\pi\|_\varrho$ measures the root-mean-square
value gap over the initial states of interest. The target density
$\varrho$ specifies where performance is measured; it need not equal
the state distribution used to collect the training batch.
The bounds below hold for every such choice of $\varrho$.
All values remain full-space discounted returns; trajectories are not
stopped at the boundary of the evaluation or sampling region.
For readability, region dependence is left implicit in the notation
for the error budgets and the final error bounds. The constants and
error-conversion map may still depend on the fixed sampling and
evaluation regions.
}

Let \(\pi_N\) be the policy returned by Algorithm~\ref{alg:phibeq_pi}
after \(N\) policy-improvement steps.  Adding and subtracting the surrogate
values gives
\begin{equation}
\label{eq:master_decomp}
\begin{aligned}
\norm{V^* - V^{\pi_N}}_{\varrho}
&\le
\underbrace{\norm{V^* - V^*_{\dt}}_{\varrho}}_{\epsilon_{\rm disc}}
+
\underbrace{\norm{V^{\pi_N}_{\dt}-V^{\pi_N}}_{\varrho}}_{\epsilon_{\rm tr}}
+
\underbrace{\norm{V^*_{\dt}-V^{\pi_N}_{\dt}}_{\varrho}}_{\epsilon_{\rm alg}} .
\end{aligned}
\end{equation}

The first two terms measure the error from replacing the continuous-time
problem by its surrogate, for the optimal value and the returned policy,
respectively.  The third measures the suboptimality of \(\pi_N\) in the
surrogate problem.  We state the common assumptions before bounding the
two transfer terms.

{\color{blue}
\begin{namedassumption}{A1 (Global dynamics)}
\label{ass:phibe-dynamics}
The state space is $\mathbb R^d$, and the action set
$\mathcal A\subset\mathbb R^m$ is compact.
\begin{enumerate}[label=(\alph*)]
\item
The reward $r$, drift $b$, and diffusion coefficient $\sigma$
admit three times continuously differentiable extensions to
$\mathbb R^d\times\mathcal U_{\mathcal A}$, where
$\mathcal U_{\mathcal A}$ is an open neighborhood of $\mathcal A$.
Writing $\Sigma:=\sigma\sigma^\top$, assume
\[
\sum_{j=0}^{3}\left(
\|\nabla_{(s,a)}^j r\|_\infty
+\|\nabla_{(s,a)}^j b\|_\infty
+\|\nabla_{(s,a)}^j\sigma\|_\infty
+\|\nabla_{(s,a)}^j\Sigma\|_\infty
\right)<\infty.
\]
Here $\nabla_{(s,a)}^j$ denotes derivatives of total order $j$ in the
state and action variables, with $\nabla_{(s,a)}^0f:=f$; the suprema
are over $\mathbb R^d\times\mathcal U_{\mathcal A}$.
\item
The covariance is independent of the action and uniformly elliptic:
\[
\Sigma(s,a):=\sigma(s,a)\sigma(s,a)^\top
=\Sigma_0(s)\succeq\lambda_0 I,
\qquad (s,a)\in\mathbb R^d\times\mathcal A,
\]
for some $\lambda_0>0$.
\end{enumerate}
\end{namedassumption}

We write
\[
r_{\max}:=\sup_{s,a}|r(s,a)|,\qquad
b_{\max}:=\sup_{s,a}\|b(s,a)\|_2,\qquad
\sigma_{\max}:=\sup_{s,a}\|\sigma(s,a)\|_{\mathrm{op}},
\]
where the suprema are over $\mathbb R^d\times\mathcal A$.

\begin{namedassumption}{A2 (Features, greedy policies, and population Gram matrix)}
\label{ass:features}
Let $\mathcal U_{\mathcal K}$ be a bounded open neighborhood of the
compact sampling region $\mathcal K$ specified in A3, and let
$\mathcal U_{\mathcal A}$ be the action neighborhood from A1.
For the greedy selectors $\pi_u$, $u\in\mathbb S_\Phi$, defined in
Section~\ref{subsec:linear_phibeq_pi}, write
$\Phi^u(s):=\Phi(s,\pi_u(s))$. More generally, write
$f^u(s):=f(s,\pi_u(s))$ and $\nabla_s^0f:=f$.
\begin{enumerate}[label=(\alph*)]
\item \textbf{Local feature regularity.}
The feature map $\Phi$ is twice continuously differentiable in the
state and action variables on
$\mathcal U_{\mathcal K}\times\mathcal U_{\mathcal A}$, with
\[
\sup_{(s,a)\in\mathcal U_{\mathcal K}\times\mathcal U_{\mathcal A}}
\|\nabla_{(s,a)}^j\Phi(s,a)\|\le B_{\Phi,j}<\infty,
\qquad j=0,1,2.
\]
Here the derivatives include mixed state--action derivatives, and
$\nabla_{(s,a)}^0\Phi:=\Phi$. Write $B_\Phi:=B_{\Phi,0}$.
\item \textbf{State regularity of the greedy policies.}
Each greedy policy $\pi_u:\mathbb R^d\to\mathcal A$ is continuously
differentiable on $\mathbb R^d$ and twice continuously differentiable
on $\mathcal U_{\mathcal K}$. There are finite constants
$B_{\pi,1},B_{\pi,2}$ such that
\[
\sup_{u\in\mathbb S_\Phi}
\|\nabla_s\pi_u\|_{L^\infty(\mathbb R^d)}\le B_{\pi,1},
\qquad
\sup_{u\in\mathbb S_\Phi}
\|\nabla_s^2\pi_u\|_{L^\infty(\mathcal U_{\mathcal K})}\le B_{\pi,2}.
\]
\item \textbf{Dependence on coefficient directions.}
For finite constants $L_j$, $j=0,1,2$,
\[
\|\nabla_s^j\Phi^u-\nabla_s^j\Phi^v\|_{L^\infty(\mathcal U_{\mathcal K})}
\le L_j\|u-v\|_2,
\qquad u,v\in\mathbb S_\Phi.
\]
\item \textbf{Population Gram matrix.}
The population Gram matrix satisfies
\[
G_\Phi:=\mathbb E_{\mu_{\mathcal K}}
[\Phi(s,a)\Phi(s,a)^\top]\succeq g_*I,
\qquad g_*>0.
\]
\end{enumerate}
\end{namedassumption}

A2(a) requires smooth features in both state and action near the
sampling region. A2(b) requires the greedy action to vary smoothly
with the state, with common derivative bounds across the policy family.
A2(c) requires small changes in the coefficient direction $u$ to
produce comparably small changes in $\Phi^u$ and its first two state
derivatives. Intuitively, (b) controls how a policy varies from state
to state, while (c) controls how the composed features vary from one
policy to another.

\tcb{Intuitively, A2(d) requires the feature functions to be linearly
independent in $L^2(\mu_{\mathcal K})$, with $g_*$ quantifying their
nondegeneracy.}

\begin{counterexample}
\label{rem:hard-greedy-scope}
Smooth features alone do not guarantee A2(c): a small change in the
coefficient direction can switch the greedy action and cause a jump
in the composed features.
Take $\Phi(s,a)=(1,a)^\top$ on $\mathcal A=[-1,1]$.
\tcb{When the action coefficient is zero, choose $\pi_u(s)=0$.
Then every greedy policy is constant in the state, so A2(b) holds
with $B_{\pi,1}=B_{\pi,2}=0$.}
For $\varepsilon>0$, the linear approximants $1\pm\varepsilon a$
have normalized coefficient directions
$u_\pm=(1,\pm\varepsilon)^\top/\sqrt{1+\varepsilon^2}$.
Their unique greedy policies are $\pi_{u_+}(s)=1$ and
$\pi_{u_-}(s)=-1$, both constant in the state.
Nevertheless, $\|\Phi^{u_+}(s)-\Phi^{u_-}(s)\|_2=2$ while
$\|u_+-u_-\|_2=2\varepsilon/\sqrt{1+\varepsilon^2}\to0$
as $\varepsilon\downarrow0$.
Thus A2(c) fails.
\end{counterexample}

\begin{namedassumption}{A3 (Sampling and local coverage)}
\label{ass:sampling}
\label{par:one-batch-sampling}
The sampled starting state--action pairs have a joint density
$\mu_{\mathcal K}$, supported in $\mathcal K\times\mathcal A$, where
$\mathcal K\subset\mathbb R^d$ is compact and
$\overline{B_{2R}}\subset\operatorname{int}(\mathcal K)$.
Successor states may lie outside $\mathcal K$.
\begin{enumerate}[label=(\alph*)]
\item \textbf{State regularity of the sampling density.}
For almost every $a\in\mathcal A$,
$\mu_{\mathcal K}(\cdot,a)\in C^2(\mathbb R^d)$, and
\[
\sum_{j=0}^2\operatorname*{ess\,sup}_{a\in\mathcal A}
\bigl\|\nabla_s^j\mu_{\mathcal K}(\cdot,a)
\bigr\|_{L^\infty(\mathbb R^d)}<\infty.
\]
\item \textbf{Local state and action coverage.}
Writing $\bar\mu_{\mathcal K}(s)
:=\int_{\mathcal A}\mu_{\mathcal K}(s,a)\,da$, there is
$m_{\mathcal K,R}>0$ such that
\[
\bar\mu_{\mathcal K}(s)\ge m_{\mathcal K,R}
\quad\text{for a.e. }s\in B_{2R}.
\tag{LSC}\label{eq:lower-state-coverage}
\]
For every $0<h\le h_A$, there is $c_h\in(0,1]$ such that
\[
\int_{\mathcal A\cap B(a_0,h)}\mu_{\mathcal K}(s,a)\,da
\ge c_h\bar\mu_{\mathcal K}(s)
\]
for a.e. $s\in B_{2R}$ and every $a_0\in\mathcal A$.
\end{enumerate}
\end{namedassumption}

\begingroup
\setlength{\intextsep}{6pt}
\begin{wrapfigure}{r}{0.46\linewidth}
\centering
\includegraphics[width=\linewidth]{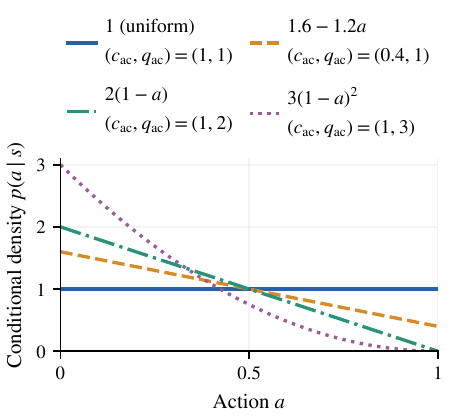}
\captionsetup{font=small,skip=3pt}
\caption{\tcb{Examples of polynomial action coverage.
Conditional action densities on $\mathcal A=[0,1]$, independent of
the state. The legend lists admissible pairs
$(c_{\rm ac},q_{\rm ac})$ for $0<h\le1$.}}
\label{fig:action-coverage}
\end{wrapfigure}
A3(a) requires the sampling density to vary smoothly with the state
and taper smoothly to zero at the boundary of its support.
\tcb{A3(b) requires the state marginal to remain uniformly positive on
$B_{2R}$, excluding gaps in state coverage. Its action-coverage condition applies to the action distribution at each sampled state. It requires every action neighborhood to have a positive probability of being sampled.}
Since $\operatorname{supp}(\varrho)\subset B_R$, performance is
evaluated within the interior of this broader sampling region.

\begin{wrappedremark}{Polynomial action coverage}
\label{rem:local-coverage}
\tcb{We consider the special case
$c_h\ge c_{\rm ac}h^{q_{\rm ac}}$ for $0<h\le h_A$, where
$c_{\rm ac},q_{\rm ac}>0$. The admissible coverage parameters depend
on the conditional action distribution and the geometry of the action
set. Figure~\ref{fig:action-coverage} illustrates
this condition using four state-independent conditional action densities
on $\mathcal A=[0,1]$, with $h_A=1$.
For these nonincreasing densities, $[1-h,1]$ is a least-sampled
radius-$h$ neighborhood; integrating each density over this interval
gives the admissible coverage parameters shown in the legend.
Compared with uniform sampling, the positive but nonuniform density
$1.6-1.2a$ retains $q_{\rm ac}=1$ with a smaller $c_{\rm ac}$.
The densities $2(1-a)$ and $3(1-a)^2$, which vanish linearly and
quadratically at the boundary, instead give $q_{\rm ac}=2$ and
$q_{\rm ac}=3$, respectively. These examples illustrate how weaker
coverage can arise through a smaller prefactor or a larger exponent.
Among the four distributions in Figure~\ref{fig:action-coverage},
uniform sampling provides the strongest worst-case action coverage.
Intuitively, poorly sampled action neighborhoods can hide large
Q-errors that mislead greedy improvement; stronger coverage limits
this effect and yields a tighter error bound, as discussed in
Remark~\ref{rem:pi-error-conversion}.}
\end{wrappedremark}
\endgroup

We next bound the two transfer errors in \eqref{eq:master_decomp},
which quantify the effect of replacing the original dynamics by the
surrogate model.

}

\begin{proposition}[Optimal-value transfer {\cite[Theorem~3.4]{ZhuZhangZhang2025OptimalPhiBE}}]
\label{prop:imported-transfer}
\tcb{Suppose \hyperref[ass:phibe-dynamics]{A1} holds. Then there is a
constant $C_{\rm opt}<\infty$ such that, for sufficiently large $\beta$
and sufficiently small $\dt$,}
\[
    \|V^* - V^*_{\Delta t}\|_\varrho
    \le C_{\rm opt}
    \left(\frac{1}{\beta^{2}}+\frac{1}{\beta^{3/2}}\right)
    \Delta t.
\]
The constant \(C_{\rm opt}\) depends only on the dynamics constants in
\hyperref[ass:phibe-dynamics]{A1}, and is independent of
\(\beta\), \(n\), \(N\), and \(\Delta t\) on the stated range.
\end{proposition}

\begin{remark}[Transfer estimate and choice of error weight]
We invoke the first-order case of
\cite[Theorem~3.4, Version~2]{ZhuZhangZhang2025OptimalPhiBE}, which
establishes a global $L^\infty$ bound. A1 verifies its
coefficient conditions through \cite[Lemma~A.1]{ZhuZhangZhang2025OptimalPhiBE},
and taking \(\beta\) sufficiently large gives the displayed dependence
on \(\beta\). Since the target $\varrho$ in
\eqref{eq:target-norm-compatibility} is a probability density,
$\|h\|_\varrho\le\|h\|_\infty$ gives the weighted estimate in
Proposition~\ref{prop:imported-transfer}.
\end{remark}

{\color{blue}
\begin{proposition}[Fixed-policy transfer]
\label{prop:fixed-policy-transfer}
Suppose \hyperref[ass:phibe-dynamics]{A1} and
\hyperref[ass:features]{A2}(a)--(b) hold. Then there is a constant
\(C_{\rm fp,tr}<\infty\) such that, for sufficiently large $\beta$
and sufficiently small $\dt$, uniformly over \(u\in\mathbb S_\Phi\),
\[
    \|V_{\dt}^{\pi_u}-V^{\pi_u}\|_\varrho
    \le \|V_{\dt}^{\pi_u}-V^{\pi_u}\|_\infty
    \le C_{\rm fp,tr}\beta^{-3/2}\dt.
\]
The thresholds and error constant depend only on the standing bounds
and fixed dimensions, and are common to the feature-induced class.
In particular, $C_{\rm fp,tr}$ is independent of $\beta,n,N,u$ and $\dt$
on the stated range.
\end{proposition}

This estimate follows from
\cite[Lemmas~A.2--A.3]{ZhuZhangZhang2025OptimalPhiBE}, applied to the
fixed-policy models. The verification of their assumptions and the
resulting discount dependence are given in
Appendix~\ref{app:fixed-policy-transfer-proof}.
}

These two estimates directly control
\(\epsilon_{\rm disc}\) and \(\epsilon_{\rm tr}\) in
\eqref{eq:master_decomp}, uniformly over the returned policy.

\subsection{Policy-evaluation error}
\label{sec:pe_error}

We next bound the error in evaluating a fixed policy \(\pi\) under the
surrogate model. To separate the effect of the feature approximation
from that of finite sampling, we introduce the population Galerkin
evaluator, obtained by replacing the empirical averages in
\eqref{eq:Ahat}--\eqref{eq:chat} with their expectations:
\begin{equation}\label{eq:population_galerkin_system}
    A^\pi_{\Delta t}\theta^\pi_{\Delta t}=c,
\end{equation}
where
\begin{equation}
\label{eq:population_galerkin_matrix}
    A^\pi_{\Delta t}:=\mathbb E[\widehat A^\pi_{\Delta t}],
    \qquad
    c:=\mathbb E[\widehat c]
      =\mathbb E_{\mu_{\mathcal K}}[r(s,a)\Phi(s,a)].
\end{equation}
Its solution defines
\(Q^\pi_{\Delta t,\Phi}=\Phi^\top\theta^\pi_{\Delta t}\).
Lemma~\ref{lem:uniform-population-stability} \tcb{ensures invertibility
of $A^\pi_{\dt}$ for sufficiently large $\beta$ and sufficiently small
$\dt$.}
This evaluator retains the same feature space as the empirical
estimator and provides an intermediate approximation between
\(Q^\pi_{\dt}\) and \(\widehat Q^\pi_{\dt,\Phi}\).
Adding and subtracting \(Q^\pi_{\dt,\Phi}\) gives
\begin{equation}\label{eq:pe-error-decomposition}
    Q^\pi_{\dt}-\widehat Q^\pi_{\dt,\Phi}
    =\underbrace{Q^\pi_{\dt}-Q^\pi_{\dt,\Phi}}_{\text{Galerkin error}}
    +\underbrace{Q^\pi_{\dt,\Phi}-\widehat Q^\pi_{\dt,\Phi}}
        _{\text{finite-sample error}}.
\end{equation}
Because the same batch also determines the policies generated during
iteration, we establish bounds uniformly over
\tcb{\(\Pi_\Phi:=\{\pi_u:u\in\mathbb S_\Phi\}\)}. We first
control the Galerkin error and then the finite-sample error.

\subsubsection{Population Galerkin approximation}
\label{sec:galerkin_error}

The Galerkin error is the difference between the exact surrogate evaluator
and its population Galerkin approximation.
\tcb{In this subsection, write $\mu=\mu_{\mathcal K}$ and
$\bar\mu=\bar\mu_{\mathcal K}$.} Define
{\color{blue}
$
    \epsilon_{\rm gal}
    :=\sup_{\pi\in\Pi_\Phi}
    \|Q^\pi_{\dt}-Q^\pi_{\dt,\Phi}\|_{L^2(\mu)}.
$
}
For a state-action function \(f\), write
\(\bar f^\pi(s)=f(s,\pi(s))\) and set
{\color{blue}
$
    \|f\|_{*,\pi}
    :=\|f\|_{L^2(\mu)}
      +\|\bar f^\pi\|_{L^2(\bar\mu)}
      +\|\bar f^\pi\|_{L^\infty(\mathcal K)}.
$
}
\begin{theorem}[Uniform Galerkin approximation]
\label{thm:uniform-galerkin-main}
\tcb{Under \hyperref[ass:phibe-dynamics]{A1}--\hyperref[ass:sampling]{A3},
there is a constant $C_{\rm gal}<\infty$ such that, for sufficiently
large $\beta$ and sufficiently small $\dt$,}
{\color{blue}
\[
    \epsilon_{\rm gal}
    \le C_{\rm gal}\sup_{\pi\in\Pi_\Phi}\inf_\theta
       \|Q^\pi_{\dt}-\Phi^\top\theta\|_{*,\pi}.
\]
}
\tcb{Here the infimum is over $\theta\in\mathbb R^{d_\phi}$.}
The constant \(C_{\rm gal}\) depends only on the standing bounds and
fixed dimensions, and is independent of \(n,N,\beta,\dt\).
\end{theorem}

The proof and an explicit bound for \(C_{\rm gal}\) are given in
Appendix~\ref{app:uniform-galerkin-proof}.

\begin{remark}
\tcb{The Galerkin error reflects the limited approximation capacity of
the chosen feature space. It vanishes if every $Q^\pi_{\dt}$,
$\pi\in\Pi_\Phi$, lies in that space. More generally, enriching the
features makes the error tend to zero if the best-approximation error
in $\|\cdot\|_{*,\pi}$ tends to zero uniformly over the policy class
and the amplification constants remain bounded.}
\end{remark}

\subsubsection{Uniform one-batch finite-sample error}
\label{sec:uniform_pe}

To control policies selected from the same batch, we bound the finite-sample
error uniformly over \(\mathbb S_\Phi\). The effect of matrix estimation
error is controlled by the common population stability margin
\[
    \alpha_\Phi(\beta,\dt)
    :=\inf_{u\in\mathbb S_\Phi}
       \sigma_{\min}(A^{\pi_u}_{\dt})
    \ge\frac{\beta g_*}{2}>0,
\]
\tcb{where $\sigma_{\min}(A)$ denotes the smallest singular value of
$A$. The lower bound follows from
Lemma~\ref{lem:uniform-population-stability} for sufficiently large
$\beta$ and sufficiently small $\dt$.}

\begin{theorem}[Uniform one-batch finite-sample error]
\label{thm:uniform-pe}
Assume \hyperref[ass:phibe-dynamics]{A1}--\hyperref[ass:sampling]{A3}.
\tcb{There are constants $C_{\rm sam},C_{\rm fs}>0$, independent of $n,\delta,\beta,\dt$, such that, for sufficiently large $\beta$,
sufficiently small $\dt$, and any $\delta\in(0,1)$, if
\mbox{$n\ge C_{\rm sam}(1+\beta^{-2}\dt^{-1})[\log(cd_\phi/\delta)+\max\{d_\phi-1,1\}\log(C_{\mathrm{net}}n)]$},}
then, with probability at least \(1-\delta\), every
\(\widehat A^{\pi_u}_{\dt}\) is invertible and, 
\[
\sup_{u\in\mathbb S_\Phi}\norm{\widehat Q^{\pi_u}_{\dt,\Phi}-Q^{\pi_u}_{\dt,\Phi}}_{L^2(\mu_{\mathcal K})}\le C_{\rm fs}(\beta^{-1}+\beta^{-2}\dt^{-1/2})\sqrt{\frac{\log(cd_\phi/\delta)+\max\{d_\phi-1,1\}\log(C_{\mathrm{net}}n)}{n}}.
\]
\tcb{Here $c,C_{\mathrm{net}}>1$ are universal constants.
The constants $C_{\rm sam}$ and $C_{\rm fs}$} depend only on the dynamics and feature bounds, the Gram
lower bound \(g_*\), and the fixed dimensions.  The sharper budget and the proof are given in
Appendix~\ref{app:sw-pe:uniform-matrix}.
\end{theorem}

\begin{remark}
\tcb{The sample condition ensures that the empirical Galerkin matrices
are invertible with high probability. For a prescribed accuracy
$\eta>0$, a sufficient condition for uniform finite-sample error at most
$\eta$, with probability at least $1-\delta$, is $n\ge C[\log(cd_\phi/\delta)+\max\{d_\phi-1,1\}\log(C_{\mathrm{net}}n)][1+\beta^{-2}\dt^{-1}+\eta^{-2}(\beta^{-2}+\beta^{-4}\dt^{-1})]$.
Here $C$ has the same allowed dependencies as the theorem constants.
When $\beta\sqrt{\dt}\le1$, the error bound reduces to
$C\beta^{-2}\sqrt{\frac{\log(cd_\phi/\delta)+\max\{d_\phi-1,1\}\log(C_{\mathrm{net}}n)}{n\dt}}$.
Thus, for a fixed discount rate, a smaller sampling interval requires a
larger batch to maintain the same finite-sample accuracy.}
\end{remark}

\begin{remark}
\tcb{For a fixed policy, the concentration bound involves
$\log(cd_\phi/\delta)$. Uniform control adds the covering term
$\max\{d_\phi-1,1\}\log(C_{\mathrm{net}}n)$} and uses the common stability
margin \(\alpha_\Phi\).  This allows every policy chosen from the batch
to be evaluated on the same event, with no additional union bound over
policy iterations. 
\end{remark}

\subsection{Policy-iteration error}
\label{sec:pi_prop}

\tcb{In this section, we bound the policy-iteration error in terms of
the Q-evaluation error.}

\begin{theorem}[Deterministic PI propagation from Q-error]
\label{thm:q-error-pi-propagation}
\label{thm:phibe-pi-contraction}
Assume \hyperref[ass:phibe-dynamics]{A1}--\tcb{\hyperref[ass:sampling]{A3}}.
{\color{blue}Fix $R\ge1$.} Let \(N\ge1\), and
let \(\{\pi_k\}_{k=0}^{N}\) be the policies generated by
Algorithm~\ref{alg:phibeq_pi}.
If, for some error budget \(\epsilon_Q\ge0\),
\[
    \sup_{0\le j<N}
    \|Q_{\dt}^{\pi_j}
        -
        \widehat Q_{\dt,\Phi}^{\pi_j}\|_{L^2(\mu_{\mathcal K})}
    \le
    \epsilon_Q,
\]
\tcb{then there is a constant $C_{\rm prop}<\infty$, independent of
\tcrev{$R,n,N,\epsilon_Q,\beta,\dt$}, such that, for sufficiently large $\beta$
and sufficiently small $\dt$,}
\[
    \|V_{\dt}^{\pi_N}-V_{\dt}^*\|_{\varrho}
    \le
    \tcb{\tcrev{C_{\rm prop}R}}
    \left[
        2^{-N/2}
        +\tcb{\Gamma}(\epsilon_Q)
        +\dt
        +\tcrev{e^{-R}}
    \right].
\]
{\color{blue}\tcrev{The constant $C_{\rm prop}$ depends on
dimension, ellipticity, the global coefficient bounds in
\hyperref[ass:phibe-dynamics]{A1}, and the global closed-loop
regularity bound derived in Lemma~\ref{lem:composed-regularity}.
The dependence on state and action coverage enters through
$\Gamma$, defined in \eqref{eq:gamma-R-map} in
Appendix~\ref{app:pi-residual-budget}. The map $\Gamma$ may still depend
on $R$ through the coverage and score-error envelope bounds.} }
\end{theorem}

The proof is given in Appendix~\ref{app:proof-main-pi}.

\begin{remark}
\label{rem:pi-error-conversion}
\tcb{The term $\Gamma(\epsilon_Q)$ controls how much Q-evaluation error
can affect greedy action selection. A small average Q-error can hide a
large overestimate at a rarely sampled action, which greedy improvement
may then select. Holding other quantities fixed, better state and action
coverage and errors that vary more slowly across actions make such
mistakes harder to hide and give a smaller bound. More accurate
Q-evaluation also reduces the bound, with $\Gamma(\epsilon_Q)\to0$
as $\epsilon_Q\to0$. Its explicit expression is given in
\eqref{eq:gamma-R-map} in Appendix~\ref{app:pi-residual-budget}.}

\tcb{Under the polynomial action-coverage condition in
Remark~\ref{rem:local-coverage} and a positive action-Lipschitz bound
for the score error, we have
\[
    \Gamma(\epsilon_Q)
    \lesssim \left(\frac{\epsilon_Q^2}{c_{\rm ac}}\right)^{1/[d(q_{\rm ac}+2)]}
\]
for sufficiently small $\epsilon_Q$, with the remaining parameters fixed.
A larger $q_{\rm ac}$, which permits weaker coverage at small action
scales, gives a smaller error exponent and hence slower decay of the
bound as $\epsilon_Q\to0$. For a fixed $q_{\rm ac}$, a larger
$c_{\rm ac}$ reduces the prefactor.}

\tcb{For the four examples in Remark~\ref{rem:local-coverage},
uniform sampling therefore gives the most favorable bound,
with the remaining quantities fixed and sufficiently small
$\epsilon_Q$. The density $1.6-1.2a$ gives the same decay
exponent but a larger prefactor, whereas $2(1-a)$ and
$3(1-a)^2$ give progressively slower decay rates.}

\tcb{For $\mathcal A=[0,1]^{d_a}$ with a conditional behavior density bounded
below, one may take $q_{\rm ac}=d_a$, giving the exponent
$2/[d(d_a+2)]$. Thus, even with a uniformly positive conditional action
density, the bound decays more slowly in higher state or action dimension.
The constant and small-error threshold are specified in
Appendix~\ref{app:pi-residual-budget}.}%

\end{remark}
\begin{remark}
\tcb{The term \tcrev{$R2^{-N/2}$} reflects the influence of the initial policy,
which decreases with further policy-improvement steps. The \tcrev{$R\dt$} term
arises from the $O(\dt)$ difference between the surrogate covariance
and the action-independent covariance of the original dynamics.
Controlling this perturbation in the PI analysis leaves a remainder
that decreases with the sampling interval.}

\tcb{\tcrev{The term $Re^{-R}$ comes from localizing the error estimate:}
we control improvement errors on $B_{2R}$ to bound the value error
on the smaller region $B_R$. The surrounding region provides a buffer
that suppresses the boundary contribution. Increasing $R$ widens this
buffer and reduces the localization term, but requires sampling coverage
over a larger region. \tcrev{The constant $C_{\rm prop}$ is independent of $R$,
but the other PI terms carry an explicit factor $R$, and $\Gamma$
may also depend on $R$. Thus the full bound need not decrease with $R$.
In particular, this bound alone does not imply convergence as $R\to\infty$.}}
\end{remark}

\subsection{Local end-to-end main theorem}
\label{subsec:end_to_end}

{\color{blue}
We combine the transfer estimates with the policy-evaluation and PI
bounds to control the returned policy in the original problem.

\begin{theorem}[Local end-to-end control guarantee]
\label{thm:end-to-end}
Assume \hyperref[ass:phibe-dynamics]{A1}--\tcb{\hyperref[ass:sampling]{A3}}
and the one-batch sampling model in \eqref{eq:def-of-data}.
Fix $R\ge1$, $N\ge1$, and $\delta\in(0,1)$.
For sufficiently large $\beta$ and sufficiently small $\dt$,
if $n$ satisfies the sample-size condition in Theorem~\ref{thm:uniform-pe},
the output $\pi_N$ of Algorithm~\ref{alg:phibeq_pi} satisfies, with probability at least
$1-\delta$,
\[
\|V^*-V^{\pi_N}\|_\varrho
\le{}C_{\rm ST}
\bigl(\beta^{-2}+\beta^{-3/2}\bigr)\dt+\tcrev{C_{\rm prop}R}\left[
2^{-N/2}+\Gamma\bigl(\epsilon_{\rm PE}\bigr)
+\dt+\tcrev{e^{-R}}\right],
\]
where $C_{\rm ST}:=C_{\rm opt}+C_{{\rm fp,tr}}$,
$\Gamma$ is defined in \eqref{eq:gamma-R-map} in
Appendix~\ref{app:pi-residual-budget}, and
\[
\epsilon_{\rm PE}\le C_{\rm gal}\sup_{\pi\in\Pi_\Phi}\inf_\theta
\|Q^\pi_{\dt}-\Phi^\top\theta\|_{*,\pi}+C_{\rm fs}\left(\frac1\beta+\frac1{\beta^2\sqrt{\dt}}\right)
\sqrt{\frac{\log(cd_\phi/\delta)+\max\{d_\phi-1,1\}\log(C_{\mathrm{net}}n)}n}.
\]
The constants have the dependencies stated in the transfer, evaluation,
and PI results. In particular, $C_{\rm ST}$ is independent of
\tcrev{$\beta,n,N$ and $\dt$, while $C_{\rm prop}$ is also independent of $R$.
The coverage and envelope quantities in $\Gamma$ may depend on $R$ and
$(n,\delta;\beta,\dt)$, but not on the realized batch or iteration index.}
Since $\Gamma$ is nondecreasing, the displayed upper bound for
$\epsilon_{\rm PE}$ may be substituted into
the control bound.
\end{theorem}

\begin{proof}
On the uniform event of Theorem~\ref{thm:uniform-pe}, which has
probability at least $1-\delta$, the decomposition
\eqref{eq:pe-error-decomposition} and
Theorem~\ref{thm:uniform-galerkin-main} bound the evaluation error
by $\epsilon_{\rm PE}$ at every iteration.
Combining this bound with
Theorem~\ref{thm:q-error-pi-propagation} and
Propositions~\ref{prop:imported-transfer}
and~\ref{prop:fixed-policy-transfer}
in \eqref{eq:master_decomp} proves the claim.
\end{proof}
}

{\color{blue}
\begin{remark}[Performance under polynomial action coverage]
\label{rem:end-to-end-poly}
We specialize our end-to-end bound to polynomial action coverage
in the regime $\beta\sqrt{\dt}\le1$.
Under suitable conditions, combining Theorem~\ref{thm:end-to-end}
with Remark~\ref{rem:pi-error-conversion} and
\eqref{eq:gamma-R-map} gives the leading-order bound
\[
\|V^*-V^{\pi_N}\|_\varrho
=\widetilde O\Bigl(\tcrev{R}\bigl[
2^{-N/2}+\dt+\tcrev{e^{-R}}+(\beta\epsilon_{\rm gal})^{2/[d(q_{\rm ac}+2)]}
+(\beta^2n\dt)^{-1/[d(q_{\rm ac}+2)]}
\bigr]\Bigr).
\]
Here $\widetilde O$ suppresses logarithmic factors, with constants
\tcrev{depending on the fixed coverage, envelope, and feature parameters.
The explicit factor $R$ comes from PI propagation; the suppressed
parameters may still depend on the sampling region and $R$.}

The bound highlights the interaction between feature approximation
and data accuracy. More expressive features can reduce
$\epsilon_{\rm gal}$, but may require more data and better control
of the Gram conditioning. Decreasing $\dt$ reduces the deterministic
error while increasing the sampling contribution at fixed $n$;
a larger batch compensates for this effect.
Increasing $\beta$ also reduces the displayed sampling contribution,
although it changes the value functions and hence the approximation
error $\epsilon_{\rm gal}$.

The exponents also show that a larger state dimension $d$ or weaker
action coverage, represented by a larger $q_{\rm ac}$, slows the
decay of the bound as evaluation accuracy or sample size improves.
For the action cube with a conditional behavior density bounded below
in Remark~\ref{rem:pi-error-conversion}, we have $q_{\rm ac}=d_a$,
so a higher action dimension has the same effect.
Intuitively, in higher dimensions or under weaker coverage,
small average evaluation errors provide less control over the
errors relevant to greedy action selection.
More policy-improvement steps reduce the initial-policy contribution,
while the remaining terms determine the accuracy supported by the
features, data, and sampling interval, as summarized in
Table~\ref{tab:error_summary}.
\end{remark}

\begin{table}[!h]
\color{blue}
\centering
\small
\setlength{\tabcolsep}{5pt}
\renewcommand{\arraystretch}{1.22}
\begin{tabularx}{\linewidth}{@{}
>{\raggedright\arraybackslash}X
>{\raggedright\arraybackslash}X
>{\raggedright\arraybackslash}X@{}}
\toprule
Where the error comes from & Size and role in the bound & How to reduce it \\
\midrule
\textbf{Replacing the dynamics.}\newline
The one-step surrogate approximates the continuous-time problem.
& \tcrev{$O(\dt)$ at fixed $\beta$ and $R$; added directly to the control error.}
& Decrease $\dt$, while increasing the batch size to control sampling error. \\
\midrule
\textbf{Using a finite feature space.}\newline
The features cannot represent every policy's Q-function exactly.
& $\epsilon_{\rm gal}$; part of the evaluation error \tcrev{before propagating into the PI error}.
& richer independent bases \\
\midrule
\textbf{Estimating from a finite batch.}\newline
The Galerkin equations are estimated from sampled transitions.
& $\epsilon_{\rm fs}$; bounded by
$\widetilde O((n\dt)^{-1/2})$; part of the evaluation error \tcrev{before propagating into the PI error}.
& Increase $n$. A smaller $\dt$ requires more samples for the same evaluation accuracy. \\
\midrule
\textbf{Stopping policy iteration.}\newline
Only $N$ greedy updates are performed.
& \tcrev{$R2^{-N/2}$; added directly, up to the $R$-independent $C_{\rm prop}$.}
& Increase $N$; evaluation and model errors remain. \\
\midrule
\textbf{Using a local residual bound.}\newline
The proof controls error on $B_R$ using residuals on $B_{2R}$.
& \tcrev{$Re^{-R}$, multiplied by the $R$-independent $C_{\rm prop}$.}
& A larger $R$ reduces this factor, but requires coverage on a larger region and control of the constants. \\
\bottomrule
\end{tabularx}
\par\smallskip
\begin{minipage}{\linewidth}
\small
\end{minipage}
\caption{Sources of error in the end-to-end guarantee and the parameters
that control them.}
\label{tab:error_summary}
\end{table}
\FloatBarrier
}

\section{Experiments}
\label{sec:experiments}

\tcb{We study three examples. First, a smooth stochastic control problem
with an analytic optimal value provides an instance satisfying
\hyperref[ass:phibe-dynamics]{A1}--\hyperref[ass:sampling]{A3}
and allows us to measure the control error of hard-greedy \tcrev{PhiBE-PI}. Second, scalar LQR illustrates the effects of the
discount parameter and sampling interval on statistical and
discretization errors. Third, CartPole examines performance when
the training and deployment frequencies differ. In the LQR and
CartPole comparisons, \tcrev{PhiBE-PI} and the \tcrev{Bellman policy iteration (BE-PI)} baseline use the same
off-policy data and features. The LQR experiment uses unbounded costs,
uniform sampling, and a drift-only evaluator; CartPole uses deterministic
rollout data, discrete actions, and softmax improvement. These two
experiments explore settings beyond the assumptions of the hard-greedy
guarantee.}

\subsection{\tcb{Smooth control with an analytic optimal value}}
\label{subsec:smooth-control-experiments}

\begin{figure}[htbp]
\centering
\includegraphics[width=\textwidth]{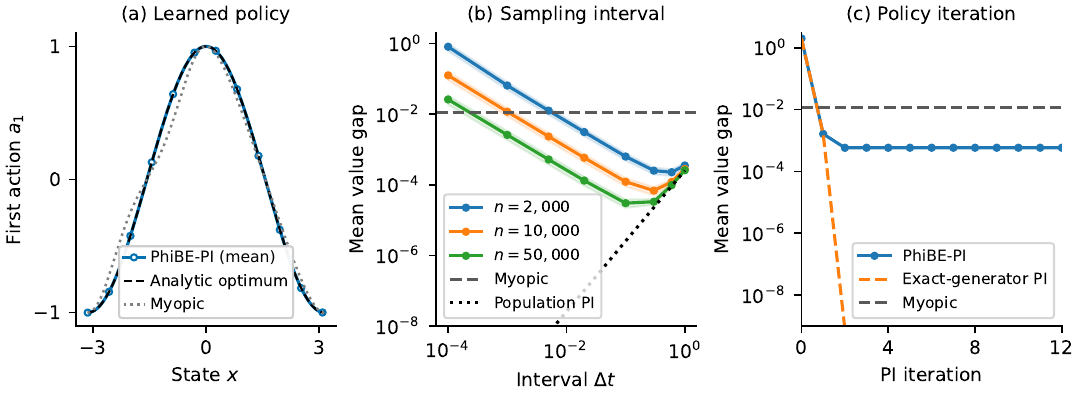}
\caption{\tcb{Smooth control with an analytic optimum.
(a) First action component: \tcrev{PhiBE-PI} mean over 100 learned policies
at $n=10000$, $\Delta t=0.02$, with pointwise $\pm1$ sample standard deviation,
compared with the analytic optimum and the myopic reference.
(b) Mean continuous-time value gap over 100 trials; shading denotes
$\pm1.96$ standard errors, and the dotted curve shows population
Galerkin PI.
(c) PI progress for $n=10000$, $\Delta t=0.02$, compared with
population PI using the exact generator; both start from $u_0=e_2$.
Horizontal dashed lines in (b)--(c) indicate the myopic gap
$J^*-J(\pi_{\mathrm{myopic}})$.}}
\label{fig:smooth-control-performance}
\end{figure}

{\bl
In a setting satisfying \hyperref[ass:phibe-dynamics]{A1}--\hyperref[ass:sampling]{A3},
we test whether data-driven \tcrev{PhiBE-PI} approaches a known analytic
optimum, and how its performance depends on the batch size, sampling
interval, and number of PI updates.

\paragraph{Setup.}
Consider \(\mathrm dX_t=a_{1,t}\,\mathrm dt+\mathrm dW_t\) on
\(\mathbb R\), with actions in the closed unit disk and \(\beta=2\).
Let \(v_0(x)=2+\sin x\), let \(O(x)\) be the rotation matrix with
angle \(x\), and set \(\Phi(x,a)=v_0(x)O(x)^\top a\) and
\(r(x,a)=2v_0(x)(a_1\cos x+a_2\sin x)-a_1\cos x+\tfrac12\sin x\).
The reward is constructed so that \(V^*(x)=v_0(x)\) and
\(\pi^*(x)=(\cos x,\sin x)\). For every \(u\in S^1\), the unique
greedy policy and composed feature satisfy \(\pi_u(x)=O(x)u\) and
\(\Phi^u(x)=v_0(x)u\), giving the uniform state regularity and
directional Lipschitz bounds in \hyperref[ass:features]{A2} on the
entire coefficient sphere.

\paragraph{Data and implementation.}
We sample states and actions independently from
\(p_X(x)=Z^{-1}e^{-1/(16-x^2)}\mathbf1_{\{|x|<4\}}\) and
\(p_A(a)=\frac{1+0.6a_1}{\pi}\mathbf1_{\{|a|\le1\}}\), where \(Z\) is a normalizing constant. The smooth state density,
positive action coverage, and Gram bound
\(G_\Phi=\mathbb E[v_0(X)^2]I_2/4\succeq I_2\), together with the
bounded smooth coefficients and constant diffusion, verify
\hyperref[ass:phibe-dynamics]{A1}--\hyperref[ass:sampling]{A3}
with \(R=3/2\) and \(\mathcal K=[-4,4]\). The theorem's remaining
quantitative sufficient conditions are not certified here.

We generate exact transitions
\(x'=x+a_1\Delta t+\sqrt{\Delta t}\,\xi\), with
\(\xi\sim\mathcal N(0,1)\), and run 20 hard-greedy PI steps from
\(u_0=e_2\), retaining the Hessian term and reusing one batch.
We test \(n\in\{2000,10000,50000\}\) and sampling intervals from
\(10^{-4}\) to \(1\), with 100 independent trials per setting.

\paragraph{Evaluation and results.}
We measure \(J(\pi)=\mathbb E_{\varrho}[V^\pi(X)]\) for
\(\varrho=\mathrm{Unif}([-1,1])\), so \(J^*=2\).
Values are computed from the fixed-policy PDE on one period, with
mesh-refinement checks. The myopic reference
\(\pi_{\mathrm{myopic}}(x)\in\arg\max_a r(x,a)\) maximizes only
the immediate reward and serves to assess the benefit of accounting
for future state evolution. Both \tcrev{PhiBE-PI} and the myopic policy are
evaluated against the same analytic optimum, using the optimality
gap \(J^*-J(\pi)\).

Figure~\ref{fig:smooth-control-performance}(a) shows that the mean
learned first action closely tracks the analytic optimum at
\(n=10000\), \(\Delta t=0.02\). Panel (b) shows decreasing value
error with more data and a bias--variance trade-off in \(\Delta t\).
At \(\Delta t=0.02\), the mean gaps \(J^*-J(\widehat\pi)\) are
\(3.12\times10^{-3}\), \(5.84\times10^{-4}\), and
\(1.32\times10^{-4}\) for the three sample sizes, compared with
\(1.13\times10^{-2}\) for the myopic policy.
For \(n=10000\), the smallest mean gap on the tested step-size grid
occurs at \(\Delta t=0.3\). Panel (c) shows that most improvement
occurs within two PI updates at \(\Delta t=0.02\).
Very small intervals can impair learning:
at \(n=2000\), \(\Delta t=10^{-4}\), the mean gap is \(0.817\),
and 34 trials retain a direction change above \(10^{-7}\) after
20 updates. All trials are included.
}

\subsection{End-to-end LQR scaling in \(\beta\) and \(\Delta t\)}
\label{subsec:lqr-e2e-experiments}

We \tcb{next} consider scalar stochastic LQR, for which the optimal feedback gain
can be computed exactly.  The dynamics and cost are
\tcb{\(\mathrm{d}s_t = (A\,s_t + B\,a_t)\,\mathrm{d}t + \sigma\,\mathrm{d}W_t\)
and \(J(\pi) = \mathbb{E}\!\int_0^\infty e^{-\beta t}(q\,s_t^2 + r\,a_t^2)\,\mathrm{d}t\),}
with \(A=0.4\) and \(B=\sigma=q=r=1\).  We use the quadratic feature basis
\(\Phi(s,a) = [1,\ s^2,\ sa,\ a^2]^\top\),
sample \((s,a)\) uniformly from \([-2,2]\times[-2,2]\), and generate \(s'\) from
the exact one-step linear-Gaussian transition induced by the continuous-time
dynamics under piecewise-constant actions.  \tcb{We compare \tcrev{PhiBE-PI} with the discrete-time}
\tcrev{BE-PI} baseline, using the same batch and features.

\tcb{In this LQR specialization, \tcrev{PhiBE-PI} uses only the first-order
feature derivative and no Hessian term. For the quadratic value function
and constant additive diffusion, the true diffusion term contributes only
a constant to policy evaluation and does not affect the population greedy
gain. The finite-step raw second moment also contains an action-dependent
$O(\Delta t)$ correction, so the drift-only evaluator is a distinct
specialization of the full evaluator.}

\paragraph{\tcb{Evaluation.}}
\tcb{Let $K^*_{\mathrm{ct}}$ and $K^*_{\mathrm{surr}}$ denote the
continuous-time and method-specific surrogate-optimal gains, computed
analytically from the corresponding discounted Riccati equations.
For each method, we run $M=200$ independent trials with
$n=2\times10^6$ samples and compute
\(E_{\mathrm{disc}}=|K^*_{\mathrm{surr}}-K^*_{\mathrm{ct}}|\),
\(E_{\mathrm{stat}}=\frac1M\sum_{j=1}^M|\widehat K_j-K^*_{\mathrm{surr}}|\),
and \(E_{\mathrm{tot}}=\frac1M\sum_{j=1}^M|\widehat K_j-K^*_{\mathrm{ct}}|\).}
\tcb{These measure the discretization floor, statistical error, and total
gain error, respectively; the total error need not equal the sum of the
other two quantities.}

\begin{figure}[!ht]
\centering
\includegraphics[width=0.90\textwidth]{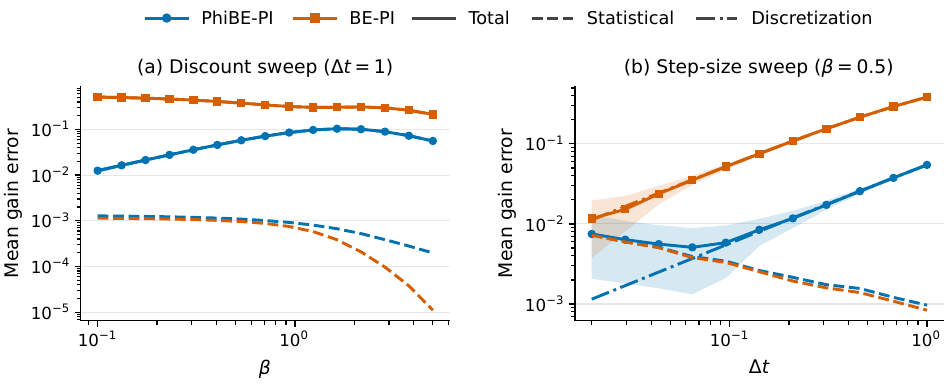}
\caption{Scalar LQR gain error, averaged over 200 trials with
$n=2\times10^6$ samples per trial.  Blue: \tcrev{PhiBE-PI}; orange: \tcrev{BE-PI}.
Solid: total error against $K^{\tcb{*}}_{\mathrm{ct}}$;
dashed: finite-sample error against $K^{\tcb{*}}_{\mathrm{surr}}$;
dash-dotted: discretization floor
$|K^{\tcb{*}}_{\mathrm{surr}}-K^{\tcb{*}}_{\mathrm{ct}}|$.
\tcb{Shaded bands around the solid curves show one standard deviation across trials.}
(a)~Varying $\beta$ at $\Delta t=1$.
(b)~Varying $\Delta t$ at $\beta=0.5$.}
\label{fig:lqr-pi-scaling}
\end{figure}

\paragraph{Results.}
Figure~\ref{fig:lqr-pi-scaling}(a) varies $\beta\in[0.1,5]$ at
$\Delta t=1$.  \tcrev{PhiBE-PI} has lower total gain error than \tcrev{BE-PI} throughout
the tested range.  \tcb{The total errors nearly coincide with the
discretization errors for both methods, indicating that discretization
dominates at the current sample size.  For \tcrev{PhiBE-PI}, the discretization
error increases over most of the range and decreases at the largest
discounts.}
\par
Figure~\ref{fig:lqr-pi-scaling}(b) fixes $\beta=0.5$\tcb{. Finite-sample error}
grows at small $\Delta t$, whereas discretization error grows at large
$\Delta t$, giving a minimum near $\Delta t=0.065$ for \tcrev{PhiBE-PI}.
\tcrev{BE-PI} has a larger discretization floor and higher total error at every
tested step size. \tcb{The explicit reference gains allow these two
sources of error to be distinguished, complementing the value-error
measurements in the preceding example.}

\subsection{\tcb{CartPole with different training and deployment frequencies}}
\label{subsec:cartpole-experiments}

\tcb{We consider CartPole control when only coarsely sampled training data
are available, but actions can be updated more frequently at deployment.
We compare \tcrev{PhiBE-PI} and \tcrev{BE-PI} using the same fixed data while
varying the deployment control interval.}

\paragraph{\tcb{Dynamics and data.}}
\tcb{The state is
$s=(x,v,\theta,\omega)$, where $x$ is cart position, $v=\dot x$,
$\theta=0$ is upright, and $\omega=\dot\theta$.
The available forces are $F_a\in\{-50,0,50\}$ N.
We use cart mass $M=8$, pole mass $m=2$, half-length $\ell=0.5$,
and gravity $g=9.8$. Writing
$c_a=(F_a+m\ell\omega^2\sin\theta)/(M+m)$, the dynamics are
\(\dot x=v\), \(\dot\theta=\omega\),
\(\dot\omega=\frac{g\sin\theta-c_a\cos\theta}
{\ell\left(4/3-\frac{m}{M+m}\cos^2\theta\right)}\), and
\(\dot v=c_a-\frac{m\ell}{M+m}\dot\omega\cos\theta\).}

\tcb{For each of 30 training seeds (9000--9029), both methods use the same
dataset of $n=8000$ action-labeled transition pairs. To collect these pairs,
we draw an initial state with independent coordinates uniform on
$[-0.2,0.2]$, then repeatedly choose a force uniformly and hold it for
$\Delta_{\mathrm{tr}}=0.1215$ s. We integrate the dynamics using
fourth-order Runge--Kutta with internal step $h=0.001215$ s and no added
force noise; deployment uses the same integrator and internal step.
A rollout terminates when $|x|\ge2.4$ m or $|\theta|\ge\pi/15$
($12^\circ$), checked after every internal step, and restarts from the same
initial-state distribution. We record $(s,a,r(s),s',\tau,d)$ for each
transition, where $\tau\le\Delta_{\mathrm{tr}}$ is the elapsed time up to
the next action update or failure, and $d$ indicates failure.
Each pair is assigned the smooth instantaneous reward
$r(s)=0.01+0.99\exp[-3(\theta/(\pi/15))^2-3(x/2.4)^2]$,
with zero value after termination.}

\paragraph{\tcb{Implementation.}}
\tcb{Both methods use 246 state--action features: an intercept and 81
Gaussian state features, tensored with three action indicators. The
Gaussians have width one and centers in $\{-1,0,1\}^4$ after scaling the
state coordinates by $(2.4,2,\pi/15,2)$. We use a common $\beta=5$
s$^{-1}$, selected from a preliminary stability scan, giving \tcrev{BE-PI} discount
$\gamma=e^{-\beta\Delta_{\mathrm{tr}}}\approx0.5447$.
Softmax logits are $Q/T$ for \tcrev{BE-PI} and $\beta Q/T$ for \tcrev{PhiBE-PI}, with $T=200$;
deployment is deterministic greedy. We add $(0.1/n)I$ to each normalized
evaluation matrix. We initialize the weights at zero and stop when their
Euclidean change is at most $10^{-5}$, with a maximum of 200 iterations.}

\tcb{Failure can shorten a transition, so policy evaluation uses its
recorded duration $\tau$. We weight each transition's contribution by
$\tau/\Delta_{\mathrm{tr}}$. \tcrev{BE-PI} discounts the next-state value by
$e^{-\beta\tau}$ and sets it to zero after failure. Its reward target is
$r(s)(1-e^{-\beta\tau})/(1-e^{-\beta\Delta_{\mathrm{tr}}})$, which equals
$r(s)$ for a full hold interval. For nonterminal transitions, \tcrev{PhiBE-PI}
approximates the rate of change of the policy-averaged value by
$\nabla\bar Q^\pi(s)\cdot(s'-s)/\tau$, differentiating both the features
and the softmax probabilities. For failed transitions, it instead
approximates this rate by $[0-\bar Q^\pi(s)]/\tau$, using zero continuation
value after failure.
Both methods use the recorded $r(s)$ without intermediate reward samples.}

\paragraph{\tcb{Evaluation.}}
\tcb{Each learned policy is evaluated on 45 paired initial states from
the same reset distribution, over a horizon $H=12.15$ s. Training is
fixed while the deployment action-update interval varies as
$\Delta_{\mathrm{ts}}=\Delta_{\mathrm{tr}}/\rho$, with
$\rho\in\{1,10,20,50,100\}$. Reward and absolute angle are averaged over
observations at action updates before termination, first within each
rollout and then equally across rollouts and training seeds. We also
report survival time, capped at $H$, and the fraction of rollouts that
fail by $H$; conditional reward alone can conceal early failure.
Shaded bands show one sample standard deviation across training seeds.}

\begin{figure}[!ht]
    \centering
    \includegraphics[width=\linewidth]{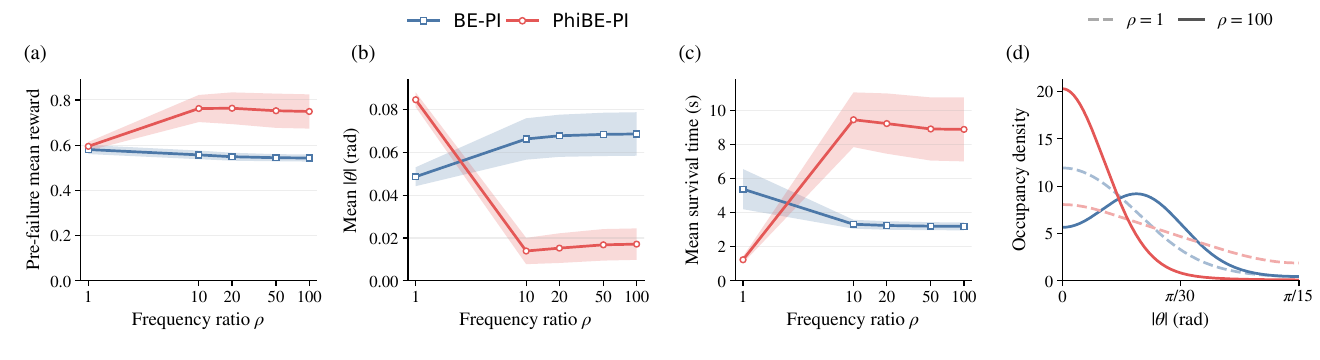}
    \caption{\tcb{CartPole with fixed training interval
    $\Delta_{\mathrm{tr}}=0.1215$ s and varying deployment frequency,
    for \tcrev{BE-PI} (blue) and \tcrev{PhiBE-PI} (red). (a) Pre-failure mean reward;
    (b) mean absolute angle; (c) mean survival time, capped at $12.15$ s.
    Curves and bands show the mean and one standard deviation across
    30 training seeds. (d) Absolute-angle occupancy for $\rho=1$
    (light dashed) and $\rho=100$ (dark solid), pooling occupied time
    from five paired rollouts per seed. Each density is normalized on
    $[0,\pi/15]$ and smoothed from 80 equal bins with a Gaussian kernel
    of bandwidth $0.035$ radians reflected at both boundaries.}}
    \label{fig:cartpole-sample-hold}
    \label{fig:cartpole-performance}
    \label{fig:cartpole-abs-angle-densities}
\end{figure}
\FloatBarrier

\paragraph{\tcb{Results.}}
\tcb{All 60 fits reach the stopping criterion within four iterations.
Figure~\ref{fig:cartpole-performance} shows that \tcrev{BE-PI} balances the pole
better at the training control interval ($\rho=1$): its mean absolute
angle is $0.0486$ radians and mean survival is $5.37$ s, compared with
$0.0846$ radians and $1.23$ s for \tcrev{PhiBE-PI}. Although \tcrev{PhiBE-PI} has a slightly
higher pre-failure reward ($0.595$ versus $0.581$), its rollouts end much
earlier.}

\tcb{At every finer control interval tested ($\rho=10,20,50,100$),
\tcrev{PhiBE-PI} outperforms \tcrev{BE-PI} in pre-failure mean reward, angular error, survival
time, and failure rate. At $\rho=100$, its pre-failure mean reward,
absolute angle, and survival
are $0.749$, $0.0171$ radians, and $8.87$ s, compared with $0.543$,
$0.0686$ radians, and $3.19$ s for \tcrev{BE-PI}. Panel~(d) also shows that \tcrev{PhiBE-PI}
spends more of its surviving time near upright at $\rho=100$ than at
$\rho=1$.}

\tcb{Thus, \tcrev{PhiBE-PI} benefits from finer control despite learning from
coarse data, whereas \tcrev{BE-PI} performs best at the training control interval.
However, increasing the control frequency does not always improve
performance: \tcrev{PhiBE-PI} has its longest mean survival at $\rho=10$
($9.44$ s), with a failure rate of $52.1\%$, compared with $61.9\%$ at
$\rho=100$.}

\FloatBarrier

\par
\section{Conclusion}

\label{sec:conclusion}

\tcb{We developed a model-free \tcrev{PhiBE-PI} framework based on PhiBE-Q
\cite{companion_ct_q} for continuous-time control from a fixed off-policy
batch. The framework learns a single state-action critic directly from
sampled increments and reuses the same data for successive policy updates.
For its linear-function-approximation implementation, our analysis connects
finite-sample policy evaluation to a local end-to-end control guarantee.
It also makes explicit the trade-off between reducing discretization error
and controlling statistical error at a fixed sample size.}

\tcb{The experiments illustrate both the benefits and limitations of this
approach. The smooth-control example measures policy performance against
an analytic optimum, while the LQR example separates discretization and
statistical errors. Both exhibit a bias--variance trade-off in the sampling
interval. In CartPole, \tcrev{PhiBE-PI} benefits from updating actions more
frequently at deployment than in the training data and outperforms the
\tcrev{BE-PI} baseline at the finer control intervals tested. The \tcrev{BE-PI}
baseline, however, performs better when the training and deployment
intervals coincide.}

\tcb{The theoretical guarantee remains local and applies to linear function
approximation with hard-greedy improvement under the stated regularity and
coverage assumptions. The softmax implementation used in CartPole lies
beyond this guarantee. Extending the analysis to soft policy updates and
nonlinear function approximation, and obtaining sharper finite-sample
bounds for repeated policy improvement, are important directions for
future work.}

\section*{Acknowledgments}
\tcb{The authors thank Helen Yuan for her help with the preliminary CartPole experiments.
C. Yao acknowledges support from the Institute for Mathematical and Statistical
Innovation (IMSI), funded by NSF grant DMS-2425650, for her participation in the
workshop ``New Directions in Reinforcement Learning and Control.'' Y. P. Zhang is partially supported by NSF CAREER grant DMS-2440215 and Simons Foundation Travel Support MPS-TSM-00007305. Y. Zhu is supported in part by the U.S. National Science Foundation under Award No.~2529107. }

\newpage
\begingroup
\normalem %
\bibliographystyle{unsrt}
\bibliography{references}
\endgroup

\newpage
\appendix
\section{Likelihood-Ratio Derivatives of Policy Averages}
\label{app:policy-average-derivatives}
\tcb{For the policy average
\(\bar Q^\pi(s)=\int_{\mathcal A}Q(s,a)\pi(a\mid s)\,da\),
suppose that \(Q\) and \(\pi\) are twice differentiable in \(s\),
that \(\pi\) is strictly positive on a state-independent support,
and that differentiation can be interchanged with integration.
Define \(g_\pi:=\nabla_s\log\pi\) and
\(H_\pi:=\nabla_s^2\log\pi\).}

\tcb{Using \(\nabla_s\pi=\pi g_\pi\) and
\(\nabla_s^2\pi=\pi(H_\pi+g_\pi g_\pi^\top)\), the product rule gives
\(\nabla_s\bar Q^\pi=\mathbb E_\pi[\nabla_sQ+Qg_\pi]\) and
\[
\nabla_s^2\bar Q^\pi
=\mathbb E_\pi\!\left[
\nabla_s^2Q+\nabla_sQ\,g_\pi^\top
+g_\pi\nabla_sQ^\top
+Q(H_\pi+g_\pi g_\pi^\top)
\right],
\]
where \(\mathbb E_\pi\) denotes expectation over
\(a\sim\pi(\cdot\mid s)\), and arguments \((s,a)\) are suppressed.}

\tcb{Evaluating these integrands at target-policy action samples gives
unbiased estimators without additional simulator transitions.
For a deterministic policy, the derivatives are those of
\(s\mapsto Q(s,\pi(s))\), wherever this composition is differentiable.}

\section{Consequences of the Standing Assumptions}
\label{app:standing-consequences}

This appendix derives the regularity and stability estimates used
in the analysis from \tcb{\hyperref[ass:phibe-dynamics]{A1}--\hyperref[ass:sampling]{A3}}.
Constants may depend on the standing bounds and fixed dimensions, but not
on the greedy direction, sample size, or sampling interval.

{\color{blue}
\subsection{Regularity of the moment coefficients}

The Galerkin analysis differentiates the moment coefficients
$b_{\dt}$ and $\Sigma_{\dt}$ when integrating by parts. We therefore
need derivative bounds that remain uniform as $\dt\to0$, despite the
factor $1/\dt$ in their definitions. The lemma below collects these
bounds, first-order consistency of the coefficients and their action
derivatives, and ellipticity. Coefficient consistency follows from
Optimal-PhiBE. We prove the joint state--action derivative bounds and
action-derivative consistency below.

\begin{lemma}[Regularity and consistency of the moment coefficients]
\label{lem:standing-regularity}
Under A1, there are $C<\infty$ and $\Delta t_{\rm reg}>0$ such that,
for every $0<\dt\le\Delta t_{\rm reg}$,
\begin{align}
\sum_{j=0}^2\sup_{s,a}\bigl(\|\nabla_{(s,a)}^j b_{\dt}\|
+\|\nabla_{(s,a)}^j\Sigma_{\dt}\|\bigr)&\le C,
\label{eq:derived-moment-bounds}\\
\sup_{s,a}\bigl(\|b_{\dt}-b\|+\|\Sigma_{\dt}-\Sigma_0\|\bigr)&\le C\dt,
\label{eq:derived-moment-consistency}\\
\Sigma_{\dt}(s,a)&\succeq\lambda_0 I/2
\end{align}
The suprema are over $\mathbb R^d\times\mathcal A$. The coefficients
are twice continuously differentiable on an action neighborhood of this
set, and the derivative bounds also hold on a fixed smaller action
neighborhood. All tensor norms are induced by Euclidean norms.
The action derivatives also satisfy
\begin{equation}\label{eq:derived-action-consistency}
\sup_{s,a}\bigl(\|\nabla_a b_{\dt}-\nabla_a b\|
+\|\nabla_a\Sigma_{\dt}-\nabla_a\Sigma\|\bigr)\le C\dt.
\end{equation}
For every fixed $R$, there are finite constants
$L_{r,R},L_{b,R},L_{\Sigma,R}$, independent of $\dt$, such that
\begin{equation}\label{eq:derived-coefficient-action-lipschitz}
\begin{aligned}
|r(s,a)-r(s,a')|&\le L_{r,R}\|a-a'\|_2,\\
\|b_{\dt}(s,a)-b_{\dt}(s,a')\|&\le L_{b,R}\|a-a'\|_2,\\
\|\Sigma_{\dt}(s,a)-\Sigma_{\dt}(s,a')\|_{\mathrm{op}}
&\le L_{\Sigma,R}\|a-a'\|_2,
\end{aligned}
\qquad s\in B_{2R},\quad a,a'\in\mathcal A.
\end{equation}
\end{lemma}
\begin{proof}
For each fixed $a\in\mathcal A$, apply the first-order case of
\cite[Lemma~A.1]{ZhuZhangZhang2025OptimalPhiBE} to the diffusion
with action held at $a$. The uniform bounds in A1 make the constants
independent of $a$, giving \eqref{eq:derived-moment-consistency}.
Since $\Sigma_0\succeq\lambda_0 I$, consistency also gives
$\Sigma_{\dt}\succeq\lambda_0 I/2$ for sufficiently small $\dt$.

Set $z=(s,a)$ and $Z_t^z=(X_t^{s,a},a)$, where $X_t^{s,a}$ starts
at $s$ with action $a$ held fixed. The augmented process has drift
$\widetilde b=(b,0)$, diffusion $\widetilde\sigma=(\sigma,0)$,
and initial condition $Z_0^z=z$. A smooth action cutoff, equal to
one near $\mathcal A$ and supported in $\mathcal U_{\mathcal A}$,
allows us to work with globally bounded $C^3$ coefficients.
Couple all initial conditions using the same Brownian motion, and
write $J_t=D_zZ_t$ and $Y_t=X_t^{s,a}-s$.
The flow is twice continuously differentiable in $L^2$.
By It\^o's formula, It\^o's isometry, and Gronwall's inequality
applied to the SDE and its first two variational equations,
\begin{equation}\label{eq:flow-fourth-moment}
\sup_z\mathbb E\|J_t\|_{\mathrm F}^4\le C,
\qquad 0\le t\le1,
\end{equation}
and
\begin{equation}\label{eq:flow-displacement-moments}
\sup_z\left(\|Y_t\|_{L^2}
+\|D_zY_t\|_{L^2}+\|D_z^2Y_t\|_{L^2}\right)
\le C\sqrt t,\qquad 0<t\le1.
\end{equation}
These bounds hold uniformly in $s$ and in a fixed neighborhood of
$\mathcal A$. The flow regularity and moment bounds also justify
differentiation under the expectations below.

The moment coefficients admit the representations
\[
b_{\dt}(s,a)=\dt^{-1}\int_0^{\dt}\mathbb E[b(Z_t)]\,dt,
\qquad
\Sigma_{\dt}(s,a)=\dt^{-1}\mathbb E[Y_{\dt}Y_{\dt}^{\top}],
\]
where $b(Z_t)=b(X_t^{s,a},a)$.
By A1, the chain and product rules, Cauchy--Schwarz, and the
preceding flow estimates,
\[
\sup_z\|D_z^j\mathbb E[b(Z_t)]\|\le C,
\qquad
\sup_z\|D_z^j\mathbb E[Y_tY_t^\top]\|\le Ct,
\qquad j=0,1,2.
\]
Substitution into the representations proves
\eqref{eq:derived-moment-bounds}. Continuity of the derivatives
follows from the flow regularity and the same moment bounds.

To prove \eqref{eq:derived-action-consistency}, write
$\mathcal L^a=b(x,a)\cdot D_x+\tfrac12\Sigma(x,a):D_x^2$,
acting componentwise with $a$ fixed, and set $g=\mathcal L^ab$.
For a fixed initial state $s$, define
$G_s(x,a)=\Sigma(x,a)+b(x,a)(x-s)^\top+(x-s)b(x,a)^\top$
and $F_s=\mathcal L^aG_s$.
Applying It\^o's formula and exchanging the time integrals gives
\begin{align*}
b_{\dt}(s,a)-b(s,a)
&=\dt^{-1}\int_0^{\dt}(\dt-t)\mathbb E[g(Z_t)]\,dt,\\
\Sigma_{\dt}(s,a)-\Sigma(s,a)
&=\dt^{-1}\int_0^{\dt}(\dt-t)\mathbb E[F_s(X_t^{s,a},a)]\,dt.
\end{align*}
By A1 and the product rule, $\|D_{(x,a)}g\|_\infty\le C$ and
$\|D_{(x,a)}F_s(x,a)\|\le C(1+|x-s|)$, with $C$ independent
of $s$. The chain rule, Cauchy--Schwarz, and the flow estimates
therefore give
\[
\sup_{s,a}\left(
\|D_a\mathbb E[g(Z_t)]\|
+\|D_a\mathbb E[F_s(X_t^{s,a},a)]\|
\right)\le C,\qquad 0<t\le1.
\]
The growth and moment bounds justify differentiation under the
expectations and time integrals. Differentiating the remainder
identities gives an $O(\dt)$ bound for each action-derivative error, proving
\eqref{eq:derived-action-consistency}.

Finally, for $f\in\{r,b_{\dt},\Sigma_{\dt}\}$, the uniform bounds
on $f$ and $D_af$ on a fixed neighborhood of the compact set
$\mathcal A$ imply \eqref{eq:derived-coefficient-action-lipschitz}:
use the mean value theorem for sufficiently close actions and
boundedness otherwise. This does not require $\mathcal A$ to be
convex. Reducing $\Delta t_{\rm reg}$ to at most $1$ completes the proof.
\end{proof}

For later use, fix $0<\Delta t_0\le\Delta t_{\rm reg}$ and
$C_\Sigma<\infty$ such that
$\sup_{s,a}\|\Sigma_{\dt}(s,a)-\Sigma_0(s)\|_{\mathrm{op}}
\le C_\Sigma\dt$ and $\Sigma_{\dt}(s,a)\succeq\lambda_0 I/2$
for $0<\dt\le\Delta t_0$.
Thus the surrogate covariance may depend on the action, with an
$O(\dt)$ deviation from $\Sigma_0$.

\subsection{Regularity of composed features and closed-loop coefficients}

We now derive the bounds needed after substituting the greedy policy.
The bounds on $\Phi^u$ will be used in the Galerkin and finite-sample
analyses; the coefficient bounds will be used in policy evaluation,
policy iteration, and fixed-policy transfer.

\begin{lemma}[Regularity of composed features and closed-loop coefficients]
\label{lem:composed-regularity}
Assume A1 and A2(a)--(b). For each $u\in\mathbb S_\Phi$, the function
$\Phi^u$ is twice continuously differentiable on $\mathcal U_{\mathcal K}$,
and
\begin{equation}\label{eq:derived-composed-feature-bounds}
\sup_u\|\nabla_s^j\Phi^u\|_{L^\infty(\mathcal U_{\mathcal K})}
\le B_j,\qquad j=0,1,2,
\end{equation}
where one may take $B_0:=B_{\Phi,0}$,
$B_1:=B_{\Phi,1}(1+B_{\pi,1})$, and
$B_2:=B_{\Phi,2}(1+B_{\pi,1})^2+B_{\Phi,1}B_{\pi,2}$.
For $f\in\{r,b,\Sigma,b_{\dt},\Sigma_{\dt}\}$, the composed function
$f^u(s)=f(s,\pi_u(s))$ is continuously differentiable on $\mathbb R^d$
and twice continuously differentiable on $\mathcal U_{\mathcal K}$.
There is a common constant $C_{\rm cl}<\infty$ such that, for
$0<\dt\le\Delta t_{\rm reg}$,
\begin{equation}\label{eq:derived-closed-loop-bounds}
\sup_u\left[
\sum_{j=0}^1\|\nabla_s^jf^u\|_{L^\infty(\mathbb R^d)}
+\|\nabla_s^2f^u\|_{L^\infty(\mathcal U_{\mathcal K})}
\right]\le C_{\rm cl}.
\end{equation}
The true and surrogate covariances are bounded below by
$\lambda_0 I/2$. Their positive square roots are globally continuously
differentiable with uniformly bounded first derivatives. Moreover,
\begin{equation}\label{eq:derived-closed-loop-consistency}
\sup_u\left(
\|b_{\dt}^u-b^u\|_\infty
+\|\Sigma_{\dt}^u-\Sigma_0\|_\infty
+\|(\Sigma_{\dt}^u)^{1/2}-\Sigma_0^{1/2}\|_\infty
\right)\le C\dt.
\end{equation}
The norms in the last display are global state sup norms. The constants
depend only on the bounds in A1 and A2(a)--(b), the fixed action
neighborhood, and dimensions, and are independent of $u,\beta,n,N,\dt$.
\end{lemma}

\begin{proof}
By the chain rule, A1, A2(a)--(b), and
Lemma~\ref{lem:standing-regularity} give the stated regularity and
the bounds \eqref{eq:derived-composed-feature-bounds} and
\eqref{eq:derived-closed-loop-bounds}.
Uniform ellipticity and consistency of the composed coefficients
follow from A1 and Lemma~\ref{lem:standing-regularity} by substituting
$a=\pi_u(s)$ and using $\Sigma(s,a)=\Sigma_0(s)$.

For $P,Q\succeq(\lambda_0/2)I$, the standard estimate for the
Sylvester equation $P^{1/2}E+EQ^{1/2}=P-Q$, with
$E=P^{1/2}-Q^{1/2}$, gives
\begin{equation}\label{eq:covariance-square-root-bound}
\|P^{1/2}-Q^{1/2}\|_{\rm F}
\le\frac{\|P-Q\|_{\rm F}}{2\sqrt{\lambda_0/2}}.
\end{equation}
Smoothness of the positive square-root map and this estimate,
together with \eqref{eq:derived-closed-loop-bounds}, give the
uniform square-root derivative bounds. Applying the same estimate
to $P=\Sigma_{\dt}^u(s)$ and $Q=\Sigma_0(s)$ proves
\eqref{eq:derived-closed-loop-consistency}.
\end{proof}

\begin{lemma}[Action regularity of the features]
\label{lem:feature-action-regularity}
Let $\mathcal A$ be compact as in A1. Under A2(a), there is
$L_{\Phi,R}<\infty$ such that
\[
\|\Phi(s,a)-\Phi(s,a')\|_2\le L_{\Phi,R}\|a-a'\|_2,
\qquad s\in B_{2R},\quad a,a'\in\mathcal A.
\]
Here $B_{2R}\subset\mathcal U_{\mathcal K}$ is the region specified
in A3. The constant is independent of $u,\beta,n,N,\dt$.
\end{lemma}
\begin{proof}
By A2(a), the same compact-neighborhood argument used at the end
of the proof of Lemma~\ref{lem:standing-regularity} gives the stated
action-Lipschitz bound.
\end{proof}
}

\subsection{Population Galerkin stability}

\begin{lemma}[Uniform population stability]
\label{lem:uniform-population-stability}
Under \hyperref[ass:phibe-dynamics]{A1}--\hyperref[ass:sampling]{A3},
there is \(C_D<\infty\), independent of \(\beta,u,\dt\), such that
\[
    \sigma_{\min}(A_u(\beta,\dt))\ge\beta g_*-C_D,
    \qquad 0<\dt\le\Delta t_{\rm reg}.
\]
In particular, with \(\beta_{\rm stab}:=\max\{1,2C_D/g_*\}\),
\(\alpha_\Phi(\beta,\dt)\ge\beta g_*/2\) for
\(\beta\ge\beta_{\rm stab}\).
\end{lemma}
\begin{proof}
Conditioning on the sampled state and action gives
\[
    A_u=\beta G_\Phi-D_{u,\dt},\qquad
    D_{u,\dt}:=\mathbb E_{\mu_{\mathcal K}}
       [\Phi(\mathcal L_{\dt}^{a}\Phi^u)^\top].
\]
\tcb{Lemmas~\ref{lem:standing-regularity} and
\ref{lem:composed-regularity} imply}
\(\sup_{u,\dt}\|D_{u,\dt}\|_{\rm op}\le C_D\).  Consequently, for every
unit vector \(v\),
\[
    \|A_uv\|_2\ge\beta\|G_\Phi v\|_2-\|D_{u,\dt}v\|_2
    \ge\beta g_*-C_D.
\]
Taking the infimum over \(v\), then over \(u\), proves the result.
No symmetry of \(A_u\) or stationarity of the sampling law is needed.
\end{proof}

\subsection{Well-posedness of policy evaluation}

\begin{proposition}[PE equations and discounted returns]
\label{prop:pe-wellposed}
\tcb{Assume A1 and A2(a)--(b)}, and take
$0<\dt\le\min\{\Delta t_0,\Delta t_{\rm reg}\}$.
For every $u\in\mathbb S_\Phi$ and $\beta>0$, each of the equations
\[
(\beta-\mathcal L^{\pi_u})V=r^u,
\qquad
(\beta-\mathcal L_{\dt}^{\pi_u})\widehat V=r^u
\quad\text{on }\mathbb R^d
\]
has a unique solution in $C^2(\mathbb R^d)\cap L^\infty(\mathbb R^d)$.
These solutions are respectively $V^{\pi_u}$ and $V_{\dt}^{\pi_u}$,
the discounted returns of the corresponding policy processes, and
\[
\|V^{\pi_u}\|_\infty\vee\|V_{\dt}^{\pi_u}\|_\infty
\le r_{\max}/\beta.
\]
Global gradient bounds for sufficiently large $\beta$ are established
in Lemma~\ref{lem:fixed-policy-gradient} below.
\end{proposition}

\begin{proof}
We apply classical elliptic solvability and discounted verification
to each fixed-policy equation. \tcb{By Lemma~\ref{lem:composed-regularity}}, the coefficients and reward are
bounded and locally H\"older continuous, and the covariance is
uniformly elliptic. Dirichlet solvability
\cite[Theorem~6.14]{gilbarg1977elliptic}, the maximum principle,
and interior Schauder estimates yield a bounded classical solution
on $\mathbb R^d$ by exhaustion with balls. Each generator determines
a unique nonexplosive diffusion law
\cite[Chapter~7]{StroockVaradhan1979}. Applying
\cite[Theorem~IV.5.1 and Corollary~IV.5.1]{fleming_soner_2006}
with a singleton control set identifies every bounded classical
solution with the corresponding discounted return; boundedness and
$\beta>0$ ensure the required terminal condition. This proves
uniqueness, and the representation gives
\[
\|V^{\pi_u}\|_\infty\vee\|V_{\dt}^{\pi_u}\|_\infty
\le\int_0^\infty e^{-\beta t}r_{\max}\,dt
=\frac{r_{\max}}{\beta}.
\]
\end{proof}

{\color{blue}
\subsection{Uniform gradients of the policy values}

The following bound is a direct consequence of
\cite[Lemma~A.2]{ZhuZhangZhang2025OptimalPhiBE}, applied to the true
and surrogate fixed-policy models. We verify that its constants can
be chosen uniformly over the policy family and the time step.

\begin{lemma}[Fixed-policy gradient bounds]
\label{lem:fixed-policy-gradient}
Assume A1 and A2(a)--(b), with $0<\dt\le\Delta t_0$ and
$\dt\le\Delta t_{\rm reg}$.
There are constants $\beta_{\rm grad}\ge1$ and $C_{\rm grad}<\infty$,
independent of $u,\dt,\beta$, such that, for $\beta\ge\beta_{\rm grad}$,
\begin{equation}\label{eq:fixed-policy-gradient-resolvent}
\sup_{u\in\mathbb S_\Phi}
\max\!\left\{\|\nabla V^{\pi_u}\|_\infty,
              \|\nabla V_{\dt}^{\pi_u}\|_\infty\right\}
\le\frac{C_{\rm grad}}{\beta}.
\end{equation}
\end{lemma}

\begin{proof}
Lemma~\ref{lem:composed-regularity} bounds the composed rewards, drifts,
covariances, and their first state derivatives uniformly in $u,\dt$.
It also supplies the common ellipticity and square-root derivative bounds.
Thus the two fixed-policy models satisfy Assumption~2(a) of
\cite{ZhuZhangZhang2025OptimalPhiBE} with common bounds. Applying
Lemma~A.2 of that reference with a singleton control set gives
\eqref{eq:fixed-policy-gradient-resolvent} for a common sufficiently
large discount threshold. Proposition~\ref{prop:pe-wellposed}
identifies the corresponding values with $V^{\pi_u}$ and
$V_{\dt}^{\pi_u}$ and ensures their classical regularity.
\end{proof}

\subsection{Fixed-policy transfer}
\label{app:fixed-policy-transfer-proof}
We verify the application of
\cite[Lemma~A.3]{ZhuZhangZhang2025OptimalPhiBE} to the true and surrogate
fixed-policy models in Proposition~\ref{prop:fixed-policy-transfer}. Set
$\Delta t_{\rm fp}:=\min\{1,\Delta t_0,\Delta t_{\rm reg}\}$ and
$\beta_{\rm fp}:=\beta_{\rm grad}$ from
Lemma~\ref{lem:fixed-policy-gradient}.
For a fixed $u$ and $0<\dt\le\Delta t_{\rm fp}$, write
$V:=V^{\pi_u}$ and $\widehat V:=V_{\dt}^{\pi_u}$.
The derived consistency estimate
\eqref{eq:derived-closed-loop-consistency} gives
\[
\varepsilon_b:=\|b_{\dt}^u-b^u\|_\infty\le c_b\dt,
\qquad
\varepsilon_B:=\sup_s
\|(\Sigma_{\dt}^u(s))^{1/2}-\Sigma_0(s)^{1/2}\|_{\rm F}
\le c_B\dt.
\]
The two models have the same reward $r^u$. Their coefficients satisfy
Assumption~2(a) and the surrogate regularity conditions of the cited
lemma uniformly in $u,\dt$, by
Lemma~\ref{lem:composed-regularity}.

We apply the cited estimate with a singleton control set and zero
reward error. With $L:=\|\nabla V\|_\infty$, it gives
\[
\|V-\widehat V\|_\infty
\le\frac{C}{\beta}
\bigl[(\sqrt L+L)\varepsilon_B+L\varepsilon_b\bigr],
\]
where $C$ depends only on the uniform coefficient bounds and the
fixed dimensions. For $\beta\ge\beta_{\rm fp}$,
Lemma~\ref{lem:fixed-policy-gradient} gives
$L\le C_{\rm grad}/\beta$. Substituting this and the coefficient
errors above yields
\[
\|V-\widehat V\|_\varrho
\le\|V-\widehat V\|_\infty
\le C'\bigl(\beta^{-3/2}+\beta^{-2}\bigr)\dt
\le C_{\rm fp,tr}\beta^{-3/2}\dt.
\]
Here the first inequality uses that $\varrho$ is a probability
density, and the last uses $\beta\ge\beta_{\rm fp}\ge1$.
All constants are independent of $\beta,u,\dt,n,N$, as required.
}

\section{Sample Complexity for Local One-Batch Policy Evaluation}\label{app:sw-pe}

We prove the fixed-policy finite-sample bound and extend it to the greedy
class using a finite net. Let $Z=(S,A,R,S')$ follow the sampling law in
\tcb{\eqref{eq:def-of-data} (with $\mu=\mu_{\mathcal K}$)},
with independent copies $z_i$, and write $X=(S,A)$ and
$\Delta S=S'-S$.

For a vector or matrix random variable $Y$, write
\[
    \|Y\|_{\psi_\kappa}
    :=
    \inf\left\{c>0:
    \mathbb E\exp\left((\|Y\|/c)^\kappa\right)\le 2
    \right\},
\]
where $\|\cdot\|$ is the Euclidean norm for vectors and the operator norm
for matrices.
The conditional norm $\|Y\|_{\psi_\kappa\mid\mathcal G}$ replaces $\mathbb E$ by
$\mathbb E[\,\cdot\mid\mathcal G]$. Constants $C$ may vary and depend
only on the fixed region and local feature bounds; dependence on
$n,\delta,\Delta t,\beta,d,d_\phi$ and the distributional scales is
displayed explicitly.

\subsection{One-sample summands for the Galerkin objects}\label{app:sw-pe:objects}

Write \(\Phi^u=\Phi^{\pi_u}\).  The empirical Galerkin system
\eqref{eq:Ahat}--\eqref{eq:chat} has the following one-sample summands.
For $u\in \mathbb S_\Phi$, define
\[
\begin{aligned}
    g_u(Z;\beta)
    &:=
    \beta\Phi(S,A)
    - \frac{\Delta S}{\Delta t}\cdot\nabla_s\Phi^u(S)
    - \frac{\Delta S\Delta S^\top}{2\Delta t}:\nabla_s^2\Phi^u(S),\\
    M_u(Z)&:=\Phi(S,A)g_u(Z;\beta)^\top,
    \qquad
    y(Z):=R\Phi(S,A).
\end{aligned}
\]

Here $\nabla_s^2\Phi^u$ is a bilinear map from
$\mathbb R^d\times\mathbb R^d$ to $\mathbb R^{d_\phi}$, measured in
operator norm. Thus \(\widehat A_u=n^{-1}\sum_iM_u(z_i)\),
\(\widehat c=n^{-1}\sum_i y(z_i)\), \(A_u=\mathbb E M_u(Z)\), and
\(c=\mathbb E y(Z)\), where \(A_u=A^{\pi_u}_{\Delta t}\).
\tcb{We use the Gram matrix $G_\Phi$ from A2(d), the feature bound
$B_\Phi$ from A2(a), the composed-feature bounds $B_1,B_2$ from
Lemma~\ref{lem:composed-regularity}, and the direction-Lipschitz constants
$L_1,L_2$ from A2(c). The feature bounds are used at starting states
in $\mathcal K$ and actions in $\mathcal A$.} All features and feature derivatives
in the summands are evaluated at the sampled starting state \(S\),
not at \(S'\). These Lipschitz bounds also make
\(u\mapsto M_u(Z)\) continuous almost surely, so the suprema below are
measurable.

\tcb{For the proof, abbreviate the logarithmic factors in
Section~\ref{sec:uniform_pe} as}
{\color{blue}
\[
    \ell_{n,\delta}:=\log(cd_\phi/\delta),
    \qquad
    \Lambda_{n,\delta}:=\ell_{n,\delta}
      +d_{\mathrm{sph}}\log(C_{\mathrm{net}}n),
    \qquad
    d_{\mathrm{sph}}:=\max\{d_\phi-1,1\},
\]
}
\tcb{where $c,C_{\mathrm{net}}>1$ are the universal constants from
Theorem~\ref{thm:uniform-pe}.}

\subsection{Fixed-policy error and sample complexity}
\label{app:sw-pe:fixed-policy}

The prediction error is controlled by the errors in the reward vector
and Galerkin matrix.  We estimate these errors separately.

Fix \(u\in\mathbb S_\Phi\), independently of the batch, and let
\(\alpha_u=\sigma_{\min}(A_u)\).
Define the reward and matrix scales
\tcb{\(K_c:=r_{\max}B_\Phi\) and}
\[
\begin{aligned}
    K_{A,\mathbb S}(\Delta t,\beta)
    :=CB_\Phi\Bigg[&
        \beta B_\Phi
        +B_1\sqrt d
        \left(
            \frac{\sigma_{\max}}{\sqrt{\Delta t}}+b_{\max}
        \right)
        +B_2d
        \left(
            \sigma_{\max}^2+b_{\max}^2\Delta t
        \right)
    \Bigg].
\end{aligned}
\]
\tcb{Define the reward and fixed-policy matrix radii}
{\color{blue}
\[
\begin{aligned}
    r_c^{\mathrm{SE}}(n,\delta)
    &:=CK_c\left[
        \sqrt{\frac{\ell_{n,\delta}}{n}}
        +\frac{\ell_{n,\delta}}{n}
    \right],\\
    r_{A,\mathrm{fix}}^{\mathrm{SE}}
    (n,\delta;\beta,\Delta t)
    &:=CK_{A,\mathbb S}(\Delta t,\beta)
    \left[
        \sqrt{\frac{\ell_{n,\delta}}{n}}
        +\frac{\ell_{n,\delta}}{n}
    \right].
\end{aligned}
\]
}

\begin{theorem}[Fixed-policy local policy-evaluation error]
\label{thm:local-fixed-pe}
Assume \hyperref[ass:phibe-dynamics]{A1}--\hyperref[ass:sampling]{A3},
\(\beta\ge\beta_{\rm stab}\), and
\(0<\Delta t\le\min\{1,\Delta t_{\rm reg}\}\).  If
\tcb{\(r_{A,\mathrm{fix}}^{\mathrm{SE}}(n,\delta;\beta,\Delta t)
\le\alpha_u(\beta,\Delta t)/2\)}, then, with probability at least \(1-\delta\), \(\widehat A_u\) is
invertible and
\[
\begin{aligned}
    \left\|
        \widehat Q^{\pi_u}_{\Delta t,\Phi}
        -Q^{\pi_u}_{\Delta t,\Phi}
    \right\|_{L^2(\mu_{\mathcal K})}
    \le
    \sqrt{\lambda_{\max}(G_\Phi)}
    \left[
        \frac{2r_c^{\mathrm{SE}}(n,\delta)}{\alpha_u(\beta,\Delta t)}
        +
        \frac{
            2\|c\|_2r_{A,\mathrm{fix}}^{\mathrm{SE}}
            (n,\delta;\beta,\Delta t)
        }{\alpha_u(\beta,\Delta t)^2}
    \right].
\end{aligned}
\]
\end{theorem}

\begin{proof}
\tcb{We follow the concentration and perturbation argument in the proof
of Theorem~4.2 of \cite[Section~6.7]{Zhu2024PhiBE}, adapted to the
present state--action Galerkin system.
Conditioning on $X=(S,A)$, the exponential-martingale estimates under
the global coefficient bounds in A1 give
\[
\begin{aligned}
\left\|\frac{\Delta S}{\Delta t}\right\|_{\psi_2\mid X}
&\le C\sqrt d\left(
\frac{\sigma_{\max}}{\sqrt{\Delta t}}+b_{\max}\right),\\
\left\|\frac{\Delta S\Delta S^\top}{\Delta t}\right\|_{\psi_1\mid X}
&\le Cd\left(\sigma_{\max}^2+b_{\max}^2\Delta t\right).
\end{aligned}
\]
Together with the local feature bounds, these imply
\[
\|y(Z)\|_2\vee\|c\|_2\le K_c,
\qquad
\sup_u\|M_u(Z)-A_u\|_{\psi_1}
\le CK_{A,\mathbb S}(\Delta t,\beta).
\]
For fixed $u$, bounded vector Bernstein and sub-exponential matrix
Bernstein \cite[Theorem~6.2 and Remark~6.3]{Tropp2012}, with failure
probability $\delta/2$ assigned to each bound, give simultaneously
with probability at least $1-\delta$,
\[
\|\widehat c-c\|_2\le r_c^{\mathrm{SE}}(n,\delta),
\qquad
\|\widehat A_u-A_u\|_{\mathrm{op}}
\le r_{A,\mathrm{fix}}^{\mathrm{SE}}(n,\delta;\beta,\Delta t).
\]}

\tcb{Write $\theta_u:=\theta^{\pi_u}_{\Delta t}=A_u^{-1}c$ and
$\widehat\theta_u:=\widehat\theta^{\pi_u}_{\Delta t}
=\widehat A_u^{-1}\widehat c$.
On this event, the assumed matrix-radius bound and Weyl's inequality
give $\sigma_{\min}(\widehat A_u)\ge\alpha_u/2$ and
$\|\widehat A_u^{-1}\|_{\mathrm{op}}\le2/\alpha_u$, where
$\alpha_u=\alpha_u(\beta,\Delta t)$.
The resolvent identity therefore yields
\begin{equation}\label{eq:stat-coefficient-perturbation}
    \|\widehat\theta_u-\theta_u\|_2
    \le
    \frac{2}{\alpha_u}\|\widehat c-c\|_2
    +
    \frac{2\|c\|_2}{\alpha_u^2}
    \|\widehat A_u-A_u\|_{\mathrm{op}}.
\end{equation}
Combining this bound with the Gram identity
\[
    \left\|
        \widehat Q^{\pi_u}_{\Delta t,\Phi}
        -Q^{\pi_u}_{\Delta t,\Phi}
    \right\|_{L^2(\mu_{\mathcal K})}^2
    =
    (\widehat\theta_u-\theta_u)^\top
    G_\Phi
    (\widehat\theta_u-\theta_u)
\]
gives the claimed prediction bound.}
\end{proof}

\subsection{Uniform extension over the greedy class}
\label{app:sw-pe:uniform-matrix}
\label{app:sw-pe:ols}

We next make the matrix bound uniform in \(u\).  The reward vector is
shared by all policies and requires no further concentration argument.

Define the directional Lipschitz envelope
\[
    G_{\mathbb S}(Z)
    :=
    \sup_{\substack{u,v\in\mathbb S_\Phi\\u\ne v}}
    \frac{\|M_u(Z)-M_v(Z)\|_{\mathrm{op}}}{\|u-v\|_2}
\]
and the deterministic scale
\[
\begin{aligned}
    K_G(\Delta t)
    :=CB_\Phi\Bigg[&
        L_1\sqrt d
        \left(
            \frac{\sigma_{\max}}{\sqrt{\Delta t}}+b_{\max}
        \right)
        +L_2d
        \left(
            \sigma_{\max}^2+b_{\max}^2\Delta t
        \right)
    \Bigg].
\end{aligned}
\]
In \(M_u-M_v\), the common \(\beta\Phi(S,A)\) term cancels.  Applying the
direction-Lipschitz bounds in \tcb{\hyperref[ass:features]{A2}(c)} to the increment
estimates established in the proof of Theorem~\ref{thm:local-fixed-pe}
gives \tcb{\(\|G_{\mathbb S}(Z)\|_{\psi_1}\le K_G(\Delta t)\), with
\(K_G(\Delta t)=O(\Delta t^{-1/2})\)}
for fixed dynamics and feature bounds and \(0<\Delta t\le1\).

\tcb{Define \(H_{n,\delta}:=1+\sqrt{\log(c/\delta)/n}+\log(c/\delta)/n\)
and set}
\[
\begin{aligned}
    r_{A,\mathbb S}^{\mathrm{SE}}
    (n,\delta;\beta,\Delta t)
    :=C\Bigg[&
        K_{A,\mathbb S}(\Delta t,\beta)
        \left(
            \sqrt{\frac{\Lambda_{n,\delta}}{n}}
            +\frac{\Lambda_{n,\delta}}{n}
        \right)
        +
        \frac{K_G(\Delta t)H_{n,\delta}}{n}
    \Bigg].
\end{aligned}
\]

There is a finite constant
\(C_{\mathcal K}>0\), depending only on the uniform dynamics and feature
bounds in \hyperref[ass:phibe-dynamics]{A1}--\hyperref[ass:features]{A2},
the Gram lower bound \(g_*\), and the fixed dimensions, but not on
\(n,\delta,\dt\), or \(\beta\).  Define
the scale $\Gamma_\beta(\dt):=\beta+\dt^{-1/2}$ and the simplified
uniform matrix radius
\[
    \overline r_{A,\mathbb S}(n,\delta;\beta,\dt)
    :=C_{\mathcal K}\Gamma_\beta(\dt)
    \left(
        \sqrt{\frac{\Lambda_{n,\delta}}{n}}
        +\frac{\Lambda_{n,\delta}}{n}
    \right).
\]

For \(0<\Delta t\le1\), the one-sample bounds give
\(r_{A,\mathbb S}^{\rm SE}\le\overline r_{A,\mathbb S}\).
This simplified radius absorbs the interpolation remainder while retaining
the dependence on the population stability margin.

\begin{proposition}[One event controls all greedy policies]
\label{prop:sw-uniform-matrix}
Assume \hyperref[ass:phibe-dynamics]{A1}--\hyperref[ass:sampling]{A3}.  Then, for every
\(\delta\in(0,1)\), with probability at least \(1-\delta\),
\tcb{the following bounds hold simultaneously:}
{\color{blue}
\[
\begin{aligned}
    \sup_{u\in\mathbb S_\Phi}
    \left\|
        \widehat A^{\pi_u}_{\Delta t}
        -A^{\pi_u}_{\Delta t}
    \right\|_{\mathrm{op}}
    &\le r_{A,\mathbb S}^{\mathrm{SE}}
    (n,\delta;\beta,\Delta t),\\
    \|\widehat c-c\|_2
    &\le r_c^{\mathrm{SE}}(n,\delta).
\end{aligned}
\]
}
\end{proposition}

\begin{proof}
Retain the reward event from the fixed-policy proof, assigning failure
probability \(\delta/3\) to each of the reward, net, and interpolation
bounds below.

For the Galerkin matrices, let \(\mathcal N_n\) be a \(1/n\)-net of
\(\mathbb S_\Phi\).  The standard sphere covering bound gives
\tcb{\(|\mathcal N_n|\le(C_{\mathrm{net}}n)^{d_{\mathrm{sph}}}\)}.
At each \(u'\in\mathcal N_n\), apply the same sub-exponential matrix
Bernstein inequality to
\(M_{u'}(z_i)-A^{\pi_{u'}}_{\Delta t}\), and then union bound over the
net.  The \(\psi_1\) estimate in the fixed-policy proof is uniform in
\(u\), and hence gives the common envelope \(K_{A,\mathbb S}\).  Thus,
with the required probability,
\[
    \max_{u'\in\mathcal N_n}
    \left\|
        \widehat A^{\pi_{u'}}_{\Delta t}
        -A^{\pi_{u'}}_{\Delta t}
    \right\|_{\mathrm{op}}
    \le
    CK_{A,\mathbb S}
    \left[
        \sqrt{\frac{\Lambda_{n,\delta}}{n}}
        +\frac{\Lambda_{n,\delta}}{n}
    \right].
\]

It remains to move from the net to the whole sphere.  Scalar Bernstein
applied to \(G_{\mathbb S}(z_i)\) gives
\tcb{\(n^{-1}\sum_{i=1}^nG_{\mathbb S}(z_i)\le CK_GH_{n,\delta}\)}.
For any \(u\in\mathbb S_\Phi\), choose \(u'\in\mathcal N_n\) with
\(\|u-u'\|_2\le1/n\).  The empirical and population Lipschitz bounds then
give
{\color{blue}
\[
    \left\|
        \widehat A^{\pi_u}_{\Delta t}
        -A^{\pi_u}_{\Delta t}
    \right\|_{\mathrm{op}}
    \le
    \left\|
        \widehat A^{\pi_{u'}}_{\Delta t}
        -A^{\pi_{u'}}_{\Delta t}
    \right\|_{\mathrm{op}}
    +C\frac{K_GH_{n,\delta}}{n}.
\]
}
The population interpolation term is bounded by
\(n^{-1}\mathbb E G_{\mathbb S}\le CK_G/n\).  Combining these bounds
with the net event proves the claim.
\end{proof}

The refined finite-sample error budget in Theorem~\ref{thm:uniform-pe} is
\[
\begin{aligned}
    \tcb{\epsilon_{\rm fs}}
    (n,\delta;\beta,\dt)
    :=C_{\mathcal K}\Bigg[&
        \frac{1}{\alpha_\Phi(\beta,\dt)}
        \left(
            \sqrt{\frac{\ell_{n,\delta}}{n}}
            +\frac{\ell_{n,\delta}}{n}
        \right)\\
        &+
        \frac{\Gamma_\beta(\dt)}
             {\alpha_\Phi(\beta,\dt)^2}
        \left(
            \sqrt{\frac{\Lambda_{n,\delta}}{n}}
            +\frac{\Lambda_{n,\delta}}{n}
        \right)
    \Bigg].
\end{aligned}
\]

\begin{proof}[Proof of Theorem~\ref{thm:uniform-pe}]
\tcb{Take $\beta\ge\beta_{\rm stab}$ and
$0<\dt\le\min\{1,\Delta t_{\rm reg}\}$.}
We first check that the sample condition ensures stability. Write
$x:=\Lambda_{n,\delta}/n$. Lemma~\ref{lem:uniform-population-stability}
and the definition of the simplified radius give
\[
 \frac{\overline r_{A,\mathbb S}}{\alpha_\Phi}
 \le C\left(1+\frac1{\beta\sqrt{\dt}}\right)(\sqrt{x}+x).
\]
For sufficiently large $C_{\rm sam}$, \tcb{the sample-size condition
in Theorem~\ref{thm:uniform-pe}} makes this ratio at most $1/2$ and
ensures $x\le1$. On the event in
Proposition~\ref{prop:sw-uniform-matrix},
\[
\sup_u\|\widehat A_u-A_u\|_{\rm op}
\le\overline r_{A,\mathbb S}\le\alpha_\Phi/2.
\]
Since $\alpha_u\ge\alpha_\Phi$, every empirical matrix is invertible.
Applying \eqref{eq:stat-coefficient-perturbation} and the Gram identity
on this event yields
\[
\begin{aligned}
 \sup_u\norm{\widehat Q^{\pi_u}_{\dt,\Phi}
       -Q^{\pi_u}_{\dt,\Phi}}_{L^2(\mu_{\mathcal K})}
 &\le \sqrt{\lambda_{\max}(G_\Phi)}
 \left(\frac{2r_c^{\rm SE}}{\alpha_\Phi}
       +\frac{2\|c\|_2\overline r_{A,\mathbb S}}{\alpha_\Phi^2}\right)\\
 &\le\tcb{\epsilon_{\rm fs}}(n,\delta;\beta,\dt),
\end{aligned}
\]
where the last inequality follows from the radius definitions after
enlarging $C_{\mathcal K}$. Substituting
$\alpha_\Phi\ge\beta g_*/2$ into this budget and using
$\ell_{n,\delta}\le\Lambda_{n,\delta}$ and
$\sqrt{x}+x\le2\sqrt{x}$ gives
\[
\tcb{\epsilon_{\rm fs}}(n,\delta;\beta,\dt)
\le C_{\rm fs}\left(\frac1\beta+\frac1{\beta^2\sqrt{\dt}}\right)
\sqrt{\frac{\Lambda_{n,\delta}}n}.
\]
All constants depend only on the dynamics and feature bounds, $g_*$,
and the fixed dimensions, as required.
\end{proof}

\section{Population Galerkin Approximation}
\label{app:galerkin}

We prove Theorem~\ref{thm:uniform-galerkin-main} using the residual
decomposition in the proof of \cite[Theorem~4.1]{Zhu2024PhiBE}, adapted
to the state--action equation and sampling measure $\mu_{\mathcal K}$.
We bound the projected residual by integration by parts, then apply the
population stability estimate from Lemma~\ref{lem:uniform-population-stability}.
We write
\(\mu=\mu_{\mathcal K}\), \(\bar\mu=\bar\mu_{\mathcal K}\), and use the
approximation norm defined in Section~\ref{sec:galerkin_error}.

We use the population system \eqref{eq:population_galerkin_system}
with the following generator representation of its matrix:
\[
    A^\pi_{\dt}=\mathbb E_\mu
       [\Phi(\beta\Phi-\mathcal L_{\dt}^a\Phi^\pi)^\top].
\]
This follows from \eqref{eq:population_galerkin_matrix} by conditioning
\eqref{eq:Ahat} on the sampled states and actions and replacing the
increments by the moments in \eqref{eq:moment_surrogates}.

\subsection{Fixed-policy Galerkin approximation}
\label{app:proof-uniform-galerkin-main}
\label{app:galerkin-ibp}

\begin{lemma}[Fixed-policy approximation]
\label{lem:weighted-galerkin-approx}
\label{lem:local-galerkin-approx}
Assume \hyperref[ass:phibe-dynamics]{A1}--\tcb{\hyperref[ass:sampling]{A3}}.
Fix \(\pi=\pi_u\) and take
\tcb{\(0<\dt\le\min\{\Delta t_0,\Delta t_{\rm reg}\}\)}. Define the
population stability margin
\tcb{\(\alpha_\pi(\beta,\dt):=\sigma_{\min}(A^\pi_{\dt})\).}
If $\alpha_\pi(\beta,\dt)>0$, then
\tcb{\(\|Q^\pi_{\dt}-Q^\pi_{\dt,\Phi}\|_{L^2(\mu)}
\le C_{\rm gal}^{\mathcal K}(\pi;\beta,\dt)
\inf_\theta\|Q^\pi_{\dt}-\Phi^\top\theta\|_{*,\pi}\),}
where $C_{\rm gal}^{\mathcal K}(\pi;\beta,\dt)$ is given explicitly
in \eqref{eq:fixed-galerkin-constant} below.
\end{lemma}

\begin{proof}
For an arbitrary coefficient \(\theta\), set
\tcb{\(\epsilon=Q^\pi_{\dt}-\Phi^\top\theta\) and
\(\omega=\theta^\pi_{\dt}-\theta\).}
Substituting \(Q^\pi_{\dt}=\Phi^\top\theta+\epsilon\) into the surrogate
equation and using \(A^\pi_{\dt}\theta^\pi_{\dt}=c\) gives
\tcb{\(A^\pi_{\dt}\omega=\epsilon_*\), where
\(\epsilon_*:=\mathbb E_\mu
[\Phi(\beta\epsilon-\mathcal L_{\dt}^a\bar\epsilon^\pi)]\).}
Cauchy--Schwarz yields
\(\|\mathbb E_\mu[\Phi\epsilon]\|_2
\le\sqrt{\lambda_{\max}(G_\Phi)}\|\epsilon\|_{L^2(\mu)}\).

To bound the generator term in $\epsilon_*$, set
\tcb{\(f:=\bar\epsilon^\pi
=\bar Q^\pi_{\dt}-(\Phi^\pi)^\top\theta\).}
\tcb{Since \(\bar Q^\pi_{\dt}=V^\pi_{\dt}\),
Proposition~\ref{prop:pe-wellposed} and
Lemma~\ref{lem:composed-regularity} imply that $f\in C^2$
on a neighborhood of $\mathcal K$.} We transfer its derivatives onto
the coefficients and sampling density by integration by parts.
Lemma~\ref{lem:standing-regularity}, A2(a), and the product rule give
\tcb{\(\sup_{0<\dt\le\Delta t_{\rm reg}}\allowbreak
\sup_{(s,a)\in\mathcal U_{\mathcal K}\times\mathcal A}\allowbreak
\bigl[
\sum_{j=0}^1\|\nabla_s^j(\Phi b_{\dt})(s,a)\|
+\sum_{j=0}^2\|\nabla_s^j(\Phi\Sigma_{\dt})(s,a)\|
\bigr]\le C<\infty\).}
By \tcb{A3(a)}, for almost every $a\in\mathcal A$,
\tcb{\(\mu(\cdot,a)\in C^2(\mathbb R^d)\) and
\(\operatorname{supp}(\nabla_s^j\mu(\cdot,a))\subseteq\mathcal K\)
for \(j=0,1,2\).}
The uniform derivative bounds in \tcb{A3(a)} therefore imply
\tcb{\(\|\nabla_s^j\mu\|_{L^1(\mathbb R^d\times\mathcal A)}
\le |\mathcal K|\,|\mathcal A|\,\|\nabla_s^j\mu\|_\infty<\infty,
j=1,2\).}
Integration by parts therefore gives
\par\noindent
{\small
\thinmuskip=1mu \medmuskip=2mu \thickmuskip=3mu
\begin{minipage}[c]{0.466\linewidth}
\refstepcounter{equation}\label{eq:local-drift-ibp}
\[
\makebox[\linewidth]{\(\displaystyle
    \tcb{\int\Phi b_{\dt}\cdot\nabla_s f\,d\mu
    =-\int f\sum_{\ell=1}^d
         \partial_{s_\ell}(\Phi b_{\dt,\ell}\mu)\,dsda,}
\)\hfill\textup{(\theequation)}}
\]
\end{minipage}\hfill
\begin{minipage}[c]{0.52\linewidth}
\refstepcounter{equation}\label{eq:local-diffusion-ibp}
\[
\makebox[\linewidth]{\(\displaystyle
    \tcb{\int\Phi\Sigma_{\dt}:\nabla_s^2f\,d\mu
    =\int f\sum_{\ell,m=1}^d
         \partial_{s_\ell s_m}^2(\Phi\Sigma_{\dt,\ell m}\mu)\,dsda.}
\)\hfill\textup{(\theequation)}}
\]
\end{minipage}}
\par
All integrals are over \(\mathbb R^d\times\mathcal A\).  A smooth cutoff
equal to one near \(\mathcal K\) extends \(f\) without changing the
integrals, and compact support removes the boundary terms.

Define
\tcb{\(\Gamma_{2,\dt}^{\mathcal K}
:=\allowbreak\|\nabla_s(\Phi b_{\dt})\|_{L^2(\mu)}
+\allowbreak\tfrac12\|\nabla_s^2(\Phi\Sigma_{\dt})\|_{L^2(\mu)}\)}
\tcb{and \(\Gamma_{\infty,\dt}^{\mathcal K}
:=\|\Phi b_{\dt}\|_\infty\|\nabla_s\mu\|_{L^1}\)
\({}+\tfrac12\bigl[
2\|\nabla_s(\Phi\Sigma_{\dt})\|_\infty\|\nabla_s\mu\|_{L^1}\)
\({}+\|\Phi\Sigma_{\dt}\|_\infty\|\nabla_s^2\mu\|_{L^1}
\bigr]\).}
The product sup norms are over \(\mathcal K\times\mathcal A\), and the
density derivative norms are over \(\mathbb R^d\times\mathcal A\).
The derivatives in \eqref{eq:local-drift-ibp}--\eqref{eq:local-diffusion-ibp}
act on all factors.  Terms with no derivative on \(\mu\) are bounded by
Cauchy--Schwarz; the remaining terms are bounded by \(\|f\|_{\infty,\mathcal K}\)
and the \(L^1\) norms of the density derivatives.  Hence
\begin{equation}\label{eq:local-generator-bound}
    \left\|\mathbb E_\mu[\Phi\mathcal L_{\dt}^a f]\right\|_2
    \le c_d\left[
       \Gamma_{2,\dt}^{\mathcal K}\|f\|_{L^2(\bar\mu)}
       +\Gamma_{\infty,\dt}^{\mathcal K}\|f\|_{\infty,\mathcal K}
       \right].
\end{equation}
Here \(c_d\ge1\) accounts for the componentwise tensor contractions and
depends only on the state dimension. The product and density bounds
established above imply
\begin{equation}\label{eq:galerkin-uniform-gamma}
C_\Gamma:=\sup_{0<\dt\le\Delta t_{\rm reg}}
\bigl(\Gamma_{2,\dt}^{\mathcal K}
      +\Gamma_{\infty,\dt}^{\mathcal K}\bigr)<\infty.
\end{equation}

Combining \eqref{eq:local-generator-bound}, with $f=\bar\epsilon^\pi$,
and the bound for $\mathbb E_\mu[\Phi\epsilon]$ gives
\begin{equation}\label{eq:local-epsilon-star-direct}
    \tcb{\|\epsilon_*\|_2
    \le\beta\sqrt{\lambda_{\max}(G_\Phi)}\|\epsilon\|_{L^2(\mu)}
       +c_d\Gamma_{2,\dt}^{\mathcal K}\|\bar\epsilon^\pi\|_{L^2(\bar\mu)}
       +c_d\Gamma_{\infty,\dt}^{\mathcal K}
           \|\bar\epsilon^\pi\|_{\infty,\mathcal K}.}
\end{equation}
Finally, \(\|\omega\|_2\le\alpha_\pi^{-1}\|\epsilon_*\|_2\) and
\(Q^\pi_{\dt}-Q^\pi_{\dt,\Phi}=\epsilon-\Phi^\top\omega\), so
\tcb{\(\|Q^\pi_{\dt}-Q^\pi_{\dt,\Phi}\|_{L^2(\mu)}
\le\|\epsilon\|_{L^2(\mu)}
+\frac{\sqrt{\lambda_{\max}(G_\Phi)}}{\alpha_\pi}
\|\epsilon_*\|_2\).}
Substituting \eqref{eq:local-epsilon-star-direct} and using the
approximation norm from Section~\ref{sec:galerkin_error} gives the
claimed estimate for each $\theta$, with
\begin{equation}\label{eq:fixed-galerkin-constant}
    C_{\rm gal}^{\mathcal K}(\pi;\beta,\dt)
    :=1+\frac{c_d\sqrt{\lambda_{\max}(G_\Phi)}}{\alpha_\pi(\beta,\dt)}
       \left[\beta\sqrt{\lambda_{\max}(G_\Phi)}
          +\Gamma_{2,\dt}^{\mathcal K}
          +\Gamma_{\infty,\dt}^{\mathcal K}\right].
\end{equation}
\tcb{Finally, choose \(\theta_*^\pi\in\argmin_\theta
\|Q^\pi_{\dt}-\Phi^\top\theta\|_{*,\pi}\).}
\tcb{Since the preceding estimate holds for every $\theta$, applying it
with $\theta=\theta_*^\pi$ gives the claimed bound.}

To make its parameter dependence explicit, Lemma~\ref{lem:uniform-population-stability}
gives, uniformly over $\pi=\pi_u$ and $0<\dt\le\Delta t_{\rm reg}$,
\tcb{\(A^\pi_{\dt}=\beta G_\Phi-D_{\pi,\dt}\),
\(\|D_{\pi,\dt}\|_{\rm op}\le C_D\), and
\(\alpha_\pi(\beta,\dt)\ge\beta g_*-C_D\),}
where $C_D$ is independent of $\beta,\pi,\dt$.
Combining this with \eqref{eq:galerkin-uniform-gamma}, for
$\beta g_*>C_D$ we obtain
\tcb{\(C_{\rm gal}^{\mathcal K}(\pi;\beta,\dt)
\le 1+\frac{c_d\sqrt{\lambda_{\max}(G_\Phi)}}{\beta g_*-C_D}
\bigl[\beta\sqrt{\lambda_{\max}(G_\Phi)}+C_\Gamma\bigr]\).}
In particular, for $\beta\ge\beta_{\rm stab}$,
\begin{equation}\label{eq:fixed-galerkin-discount-bound}
C_{\rm gal}^{\mathcal K}(\pi;\beta,\dt)
\le 1+\frac{2c_d\lambda_{\max}(G_\Phi)}{g_*}
+\frac{2c_d\sqrt{\lambda_{\max}(G_\Phi)}C_\Gamma}{\beta g_*}.
\end{equation}
Thus the amplification factor is bounded uniformly in the time step
and has an upper bound of the form $C_0+C_1/\beta$ above the stability
threshold. The best-approximation error itself may still depend on
$\beta$ and $\dt$.
\end{proof}

\subsection{Uniform Galerkin approximation}
\label{app:uniform-galerkin-proof}

We now obtain Theorem~\ref{thm:uniform-galerkin-main} by taking the
supremum over the greedy policy class.

\begin{proof}
\tcb{Choose $\Delta t_{\rm Gal}:=\min\{\Delta t_0,\Delta t_{\rm reg}\}$
so that Proposition~\ref{prop:pe-wellposed} and
Lemma~\ref{lem:composed-regularity} apply.}
The right-hand side of \eqref{eq:fixed-galerkin-discount-bound}
is common to all $\pi_u$. For $\beta\ge\beta_{\rm stab}$, it is
bounded by
\[
    C_{\rm gal}
    :=1+\frac{2c_d\lambda_{\max}(G_\Phi)}{g_*}
       +\frac{2c_d\sqrt{\lambda_{\max}(G_\Phi)}C_\Gamma}
              {\beta_{\rm stab}g_*}.
\]
\tcb{Applying Lemma~\ref{lem:local-galerkin-approx} to each
$\pi\in\Pi_\Phi$ and taking the supremum over $\Pi_\Phi$ therefore gives
the bound in Theorem~\ref{thm:uniform-galerkin-main}.}
The constant is independent of $n,N,\beta,\dt$, as claimed.
\end{proof}

The two coefficient quantities separate the terms created by integration
by parts: \(\Gamma_{2,\dt}^{\mathcal K}\) contains derivatives of the
feature--coefficient products, while \(\Gamma_{\infty,\dt}^{\mathcal K}\)
contains derivatives of the sampling density.  The Gram matrix converts
coefficient error into prediction error, and \(\alpha_\Phi\) controls the
amplification from solving the Galerkin system.

\section{Policy-iteration propagation}
\label{app:abstract_PI}

{\color{blue}
We prove the PI propagation result in Section~\ref{sec:pi_prop}.
We first establish value regularity and contraction under a local
residual budget, then derive this budget from the Q-evaluation error.
}

\subsection{PI setup and regularity}
\label{app:pi-setup}

{\color{blue}
We work with the first-order surrogate
\[
(\widetilde r,\widetilde b,\widetilde\Sigma)
=(r,b_{\dt},\Sigma_{\dt}),
\qquad \widetilde V^\pi:=V_{\dt}^\pi.
\]
Let $\{\pi_k\}_{k=0}^N\subset\Pi_\Phi$ be the algorithmic path and
write $V^k:=\widetilde V^{\pi_k}$ and
$V^*:=V_{\dt}^*$. The initialization and the rule retaining the
previous policy when the fitted coefficient is zero keep the path in
$\Pi_\Phi$.

\subsubsection{Control equations and coefficient conditions}

For $f\in\{\widetilde r,\widetilde b,\widetilde\Sigma\}$, write
$f^\pi(s):=f(s,\pi(s))$. By
Proposition~\ref{prop:pe-wellposed}, each fixed-policy value satisfies
\[
\beta\widetilde V^\pi
=\widetilde r^\pi+\widetilde b^\pi\cdot\nabla\widetilde V^\pi
+\tfrac12\widetilde\Sigma^\pi:D^2\widetilde V^\pi.
\]
The HJB equation associated with the surrogate is
\[
\beta v(s)=\sup_{a\in\mathcal A}
\left\{\widetilde r(s,a)+\widetilde b(s,a)\cdot\nabla v(s)
+\tfrac12\widetilde\Sigma(s,a):D^2v(s)\right\}.
\]
Lemma~\ref{lem:pi-optimal-value-regularity} identifies its bounded
classical solution with the optimal value and supplies the derivative
bounds used below.

Lemma~\ref{lem:standing-regularity} gives the coefficient bounds used
below. In particular,
\[
\widetilde\Sigma=\Sigma_0+\widetilde e,\qquad
\|\widetilde e\|_\infty\le C_e\dt,\qquad
\Sigma_0\succeq\lambda_0 I.
\]
For $0<\dt\le\Delta t_{\rm reg}$,
$\widetilde\Sigma\succeq\lambda_0 I/2$. We write
$\sigma^2:=\lambda_0$ for the ellipticity constant in the PI proof.
}

{\color{blue}
\subsubsection{Value derivative bounds}
\label{app:pi-value-regularity}

Proposition~\ref{prop:pe-wellposed} supplies the bounded classical
fixed-policy values. We now bound their derivatives with constants
independent of the discount, and then establish the corresponding
bounds for the optimal value.

\begin{lemma}[Discount-uniform fixed-policy derivative bounds]
\label{lem:pi-fixed-policy-regularity}
Assume \hyperref[ass:phibe-dynamics]{A1} and
\hyperref[ass:features]{A2}(a)--(b). There is a constant $C$, independent
of $\beta,u$ and $\dt$, such that, for $\beta\ge1$ and
$0<\dt\le\Delta t_{\rm reg}$,
\begin{equation}\label{eq:pi-discount-uniform-derivatives}
\sup_u\|\nabla V_{\dt}^{\pi_u}\|_\infty\le C\beta^{-1/2},
\qquad
\sup_u\|D^2V_{\dt}^{\pi_u}\|_\infty\le C.
\end{equation}
Thus the local gradient and Hessian bounds on $B_{2R}$ may be chosen as
$G_R(\beta):=C\beta^{-1/2}$ and $H_R(\beta):=C$.
For $\beta\ge\beta_{\rm grad}$,
Lemma~\ref{lem:fixed-policy-gradient} sharpens the gradient bound to
$C_{\rm grad}/\beta$.
\end{lemma}

\begin{proof}
Fix $\beta\ge1$, a policy $\pi_u$, and a point $x\in\mathbb R^d$.
We first rescale the policy equation so that its discount coefficient
is one. Set $\rho=\beta^{-1/2}$ and
$v(y)=V_{\dt}^{\pi_u}(x+\rho y)$. Its equation on $B_1$ is
\[
\tfrac12\widetilde\Sigma^{\pi_u}(x+\rho y):D_y^2v
+\rho\widetilde b^{\pi_u}(x+\rho y)\cdot\nabla_yv-v
=-\beta^{-1}r^{\pi_u}(x+\rho y).
\]
By Lemma~\ref{lem:composed-regularity} and $\rho\le1$, the
rescaled coefficients have uniform ellipticity and $C^\alpha(B_1)$
bounds for any fixed $\alpha\in(0,1)$. Proposition~\ref{prop:pe-wellposed}
gives $\|v\|_\infty\le r_{\max}/\beta$. The interior Schauder estimate
\cite[Chapter~6]{gilbarg1977elliptic} therefore yields
\[
\|v\|_{C^{2,\alpha}(B_{1/2})}
\le C\left(\|v\|_{L^\infty(B_1)}
+\beta^{-1}\|r^{\pi_u}(x+\rho\,\cdot)\|_{C^\alpha(B_1)}\right)
\le C\beta^{-1},
\]
with $C$ independent of $\beta,u,x$ and $\dt$.
We then rescale the derivatives at $y=0$:
\[
|\nabla V_{\dt}^{\pi_u}(x)|
=\rho^{-1}|\nabla_yv(0)|\le C\beta^{-1/2},
\qquad
\|D^2V_{\dt}^{\pi_u}(x)\|
=\rho^{-2}\|D_y^2v(0)\|\le C.
\]
Taking the supremum over $u$ and $x$ proves
\eqref{eq:pi-discount-uniform-derivatives}.
\end{proof}

\begin{lemma}[Optimal-value regularity]
\label{lem:pi-optimal-value-regularity}
Under \hyperref[ass:phibe-dynamics]{A1}, for
$0<\dt\le\Delta t_{\rm reg}$ and $\beta\ge1$, the surrogate optimal
value is a bounded classical solution of the HJB equation. Moreover,
\begin{equation}\label{eq:pi-optimal-uniform-derivatives}
\|V^*\|_\infty\le r_{\max}/\beta,
\qquad \|\nabla V^*\|_\infty\le C\beta^{-1/2},
\qquad \|D^2V^*\|_\infty\le C,
\end{equation}
where $C$ is independent of $\beta$ and $\dt$.
\end{lemma}

\begin{proof}
Write $A=\widetilde\Sigma/2$. Lemma~\ref{lem:standing-regularity}
gives uniform ellipticity and uniform $C_b^1$ bounds on
$A,\widetilde b,r$, over all actions and small $\dt$.
The positive square root of $\widetilde\Sigma$ also has a uniform
Lipschitz bound, by the square-root inequality in
\eqref{eq:covariance-square-root-bound}. Thus the surrogate satisfies
Assumption~1 of \cite{ZhuZhangZhang2025OptimalPhiBE}. The value
characterization recalled in Section~2.1 of that reference identifies
$V^*$ as the unique bounded viscosity solution of its HJB equation,
and bounded rewards give $\|V^*\|_\infty\le r_{\max}/\beta$.

We first obtain classical regularity from the elliptic estimates of
\cite[Theorems~3.1, 5.1]{Safonov1994Boundary}. For
\[
F(s,z,p,X)=\sup_a\{A(s,a):X+\widetilde b(s,a)\cdot p+r(s,a)\}-\beta z,
\]
the operator is convex in $X$, uniformly elliptic, decreasing in $z$,
Lipschitz in $(z,p)$, and satisfies
\[
[F(\cdot,z,p,X)]_{\alpha;B_1(x)}
\le C(1+|p|+\|X\|).
\]
The cited Dirichlet solvability and interior estimates, applied on
expanding balls with zero boundary data and the common bound
$\|v_L\|_\infty\le r_{\max}/\beta$, give a bounded
$C^{2,\alpha}_{\rm loc}$ solution on $\mathbb R^d$ for some
$\alpha\in(0,1)$. By the viscosity uniqueness above, this solution
is $V^*$.

The cited estimates are stated for $d\ge2$. For $d=1$, apply them to
the equation on $\mathbb R^2$
\[
\sup_a\{A(s,a)v_{ss}+v_{yy}
+\widetilde b(s,a)v_s+r(s,a)\}-\beta v=0.
\]
Uniqueness and translation invariance in $y$ make the bounded solution
independent of $y$, yielding the same regularity in the original variable.

We then make the derivative estimates independent of $\beta$.
Fix $x$, set $\rho=\beta^{-1/2}$ and
$U(y)=\beta V^*(x+\rho y)$. Then
\[
\sup_a\{A(x+\rho y,a):D^2U
+\rho\widetilde b(x+\rho y,a)\cdot\nabla U
+r(x+\rho y,a)\}-U=0.
\]
Since $\rho\le1$ and the zeroth-order coefficient is $-1$,
the rescaled coefficient bounds are independent of $x,\beta$ and
$\dt$. The interior estimate and $\|U\|_\infty\le r_{\max}$ give
\[
\|U\|_{C^{2,\alpha}(B_{1/2})}\le C.
\]
Rescaling at $y=0$ gives
\[
|\nabla V^*(x)|
=\beta^{-1}\rho^{-1}|\nabla U(0)|\le C\beta^{-1/2},
\qquad
\|D^2V^*(x)\|
=\beta^{-1}\rho^{-2}\|D^2U(0)\|\le C.
\]
Taking the supremum over $x$ proves
\eqref{eq:pi-optimal-uniform-derivatives}. For sufficiently large
$\beta$, \cite[Lemma~A.2]{ZhuZhangZhang2025OptimalPhiBE} further gives
$\|\nabla V^*\|_\infty\le C/\beta$, with a common threshold and
constant because the surrogate coefficient bounds are uniform in $\dt$.
\end{proof}

\subsubsection{Greedy improvement and score residual}

Define
\[
\widetilde p^\pi(s,a)
:=\widetilde r(s,a)+\widetilde b(s,a)\cdot\nabla\widetilde V^\pi(s)
+\tfrac12\widetilde e(s,a):D^2\widetilde V^\pi(s).
\]
For the fitted evaluator $\widehat Q^\pi:=\widehat Q_{\dt,\Phi}^\pi$,
set
\[
\widehat p^\pi(s,a)
:=\beta\widehat Q^\pi(s,a)-\tfrac12\Sigma_0(s):D^2\widetilde V^\pi(s).
\]
The normalized Q identity gives
\[
\widetilde p^\pi
=\beta Q_{\dt}^\pi-\tfrac12\Sigma_0:D^2\widetilde V^\pi,
\qquad
\widetilde p^\pi-\widehat p^\pi
=\beta(Q_{\dt}^\pi-\widehat Q^\pi).
\]
The common Hessian term is independent of the action, so
\begin{equation}\label{eq:hard-greedy}
\pi_{k+1}(s)\in\arg\max_{a\in\mathcal A}\widehat p^{\pi_k}(s,a)
=\arg\max_{a\in\mathcal A}\widehat Q^{\pi_k}(s,a).
\end{equation}
The algorithm therefore does not compute this Hessian term; it is used
only to compare scores in the proof. For $1\le k\le N$, define
\begin{equation}\label{eq:abstract-approx-error}
\Delta_k(s):=\max_{a\in\mathcal A}
|\widetilde p^{\pi_{k-1}}(s,a)-\widehat p^{\pi_{k-1}}(s,a)|
=\beta\max_{a\in\mathcal A}
|Q_{\dt}^{\pi_{k-1}}(s,a)-\widehat Q^{\pi_{k-1}}(s,a)|.
\end{equation}
}

\subsection{PI contraction from a local residual budget}
\label{app:abstract-pi}

{\color{blue}We now bound the value error on $B_R$ in terms of
the score residual on $B_{2R}$.}

{\color{blue}
\begin{theorem}[Abstract PI error contraction]
\label{thm:abstract-pi}
Assume \hyperref[ass:phibe-dynamics]{A1} and
\hyperref[ass:features]{A2}(a)--(b). Let $\{\pi_k\}_{k=0}^N\subset\Pi_\Phi$
be the approximate PI path, with $V^k=V_{\dt}^{\pi_k}$ and
$0<\dt\le\Delta t_{\rm reg}$. Fix $R\ge1$.
\tcrev{There exist $C_0\ge1$ and $C>0$, depending only on the standing
coefficient and regularity bounds and independent of $R$, such that for}
$\beta\ge C_0$ and $\gamma\ge0$, the residual bound
\begin{equation}\label{eq:deltasmall_abstract}
\int_{B_{2R}}\Delta_k(s)^{2d}\,ds \;\le\; \gamma^{2d} \quad \text{  for }1\le k\le N,
\end{equation}
implies, for $1\le k\le N$,
\[
{\color{blue}\sup_{s\in B_R}|V^*(s)-V^k(s)|^2\le \tcrev{CR^2(2^{-k}+\gamma^2+\dt^2+e^{-2R})}.}
\]
\end{theorem}

{\color{blue}Lemmas~\ref{lem:pi-fixed-policy-regularity} and
\ref{lem:pi-optimal-value-regularity} supply the value and derivative
bounds used below, uniformly for $\beta\ge1$ and small $\dt$.}

\begin{proof}[Proof of Theorem~\ref{thm:abstract-pi}]
{\color{blue}
Following the PI argument of \cite{GuoTranZhang2025}, we derive a
weighted error estimate and apply the ABP estimate. Suppress tildes and set $w^k=V^*-V^k$.
Throughout the proof, $C$ depends only on the coefficient bounds,
ellipticity, and the derivative bounds in
Lemmas~\ref{lem:pi-fixed-policy-regularity} and
\ref{lem:pi-optimal-value-regularity}; it is independent of
$k,\beta\ge1,\dt$ and $R\ge1$.
Subtracting the policy equation from the HJB equation gives
}
\beq\label{7.4}
\begin{aligned}
\beta  w^k (s)-\frac{1}{2} \Sigma_0(s) :\nabla^2   w^k & =\max_{a } \left\{ r(s,a)+b(s,a)\cdot {\nabla}  V^*+\frac12e(s,a): D^2  V^*\right\}\\
&-{r}^{k}(s)-{b}^{k}(s)\cdot \nabla V^k-\frac12e^k(s): D^2  V^k.
\end{aligned}
\eeq
{\color{blue}Write $X_k$ for the right-hand side.
Choosing $a=\pi_k(s)$ gives}
{\color{blue}
\begin{align*}
X_k
&\geq b^k(s)\cdot\nabla w^k(s)+\frac12 e^k(s):D^2w^k(s)\\
&\geq -\|b\|_\infty|\nabla w^k|
-\frac12\|e\|_\infty\bigl(|D^2V^*|+|D^2V^k|\bigr)\\
&\geq -\|b\|_\infty|\nabla w^k|-C\dt.
\end{align*}
Here we used $\|e\|_\infty\le C_e\dt$ and the uniform
Hessian bounds.
}

{\color{blue}Choose}
\[
 \pi^*(s)\in\argmax_{a } \left\{ r(s,a)+b(s,a)\cdot {\nabla}  V^*+\frac12e(s,a): D^2  V^*\right\},
\]
{\color{blue}and write}
\[
p^{k}(s,a):=p^{\pi_k}(s,a)\quad \text{and}\quad\hat p^{k}(s,a):=\hat p^{\pi_k}(s,a).
\]
It follows from \eqref{eq:abstract-approx-error} and \eqref{eq:hard-greedy} that
\begin{align*}
p^{k-1}(s,\pi_k)&
\geq \hat p^{k-1}(s,\pi_k)-\Delta_k(s)\geq \hat p^{k-1}(s,\pi^*)-\Delta_k(s)\geq p^{k-1}(s,\pi^*)-2\Delta_k(s).
\end{align*}
Consequently,
\[
\begin{aligned}
 &{r}^{k}(s)+ {b}^{k}(s)\cdot \nabla V^k+\frac12 {e}^{k}(s) : D^2 V^k\\
 &\qquad=p^{k-1}(s,\pi_k)+{b}^{k}(s)\cdot (\nabla V^k-\nabla V^{k-1})+\frac12 {e}^{k}(s) : (D^2 V^k-D^2V^{k-1})\\
&\qquad \geq p^{k-1}(s,\pi^*)-2\Delta_k(s)-\|b\|_\infty(|\nabla  w^k |+|\nabla  w^{k-1} |)-C\dt,
\end{aligned}
\]
{\color{blue}The Hessian bounds control the last term. Substituting
this inequality into \eqref{7.4} gives
\[
X_k\le 2\Delta_k+2\|b\|_\infty
\bigl(|\nabla w^k|+|\nabla w^{k-1}|\bigr)+C\dt.
\]
Combining the upper and lower bounds, we obtain}

\beq\label{7.7}
|X_{k}(s)|\leq 2\Delta_k(s)+C(|\nabla  w^k |+|\nabla  w^{k-1} |+\dt).
\eeq

{\color{blue}
To localize the estimate, choose a smooth nondecreasing function $\eta:\mathbb R\to[0,1]$
with $\eta(t)=0$ for $t\le0$ and $\eta(t)=1$ for $t\ge1$.
Set
\[
\chi(t):=\int_0^t\eta(u)\,du,
\qquad \phi(s):=\exp\bigl(-2\chi(|s|-R)\bigr).
\]
Then $\phi$ is smooth, $0<\phi\le1$, and $\phi=1$ on $B_R$.
Since $\chi(t)\ge t-1$ for $t\ge1$,
\[
\sup_{\partial B_{2R}}\phi\le e^2e^{-2R}.
\]
Moreover, $|\chi'|\le1$ and $|\chi''|\le\|\eta'\|_\infty$.
The radial derivatives occur only where $|s|\ge R\ge1$, so
\[
|\nabla\phi|+\|D^2\phi\|_{\mathrm{op}}\le C\phi,
\]
with $C$ independent of $R$.
}

Multiply \eqref{7.4} by $w^k\phi$; using \eqref{7.7} we obtain
\beq\label{2.9}
\beta | w^k |^2\phi-\frac{1}{2} (\Sigma_0 : \nabla^2   w^k )  w^k \phi\leq (2\Delta_k+C (|\nabla  w^{k-1} |+|\nabla  w^{k}|+\dt)) |w^k| \phi.
\eeq
{\color{blue}Recall $\sigma^2=\lambda_0$. Young's inequality gives}
\begin{align*}
\left(2\Delta_k+C( |\nabla  w^{k-1} |+|\nabla  w^{k}|+\dt)\right)| w^k |\phi    \leq \frac{\sigma^2}{8}  (|\nabla w^{k}|^2+ |\nabla w^{k-1} |^2)\phi+\dt^2\phi+\Delta_k^2\phi +C | w^k |^2\phi.
\end{align*}

{\color{blue}The product rule gives}
\[
(\Sigma_0 : \nabla^2 w^k) w^k \phi = \frac{1}{2} \Sigma_0 : \nabla^2 ( (w^k)^2 \phi ) - (\nabla w^k)^T \Sigma_0 \nabla w^k \phi - 2 w^k (\nabla w^k)^T \Sigma_0 \nabla \phi - \frac{1}{2} (w^k)^2 (\Sigma_0 : \nabla^2 \phi).
\]
{\color{blue}Using the boundedness and ellipticity of $\Sigma_0$ and the derivative bounds for $\phi$,}
\begin{align*}
-\frac12 (\Sigma_0 :\nabla^2   w^k )  w^k \phi &\geq \frac{\sigma^2}{2} |\nabla  w^k |^2\phi - \frac{1}{4} \Sigma_0 : \nabla^2 ( |w^k|^2 \phi )-C | w^k ||\nabla  w^k |\phi-C | w^k |^2\phi\\
&\geq \frac{3\sigma^2}8 |\nabla  w^k |^2\phi - \frac{1}{4} \Sigma_0 : \nabla^2 ( |w^k|^2 \phi ) -C | w^k |^2\phi,
\end{align*}
{\color{blue}Substituting these bounds into \eqref{2.9} yields}
\begin{align*}
\beta | w^k |^2\phi - \frac{1}{4} \Sigma_0 : \nabla^2 ( |w^k|^2 \phi ) +\frac{\sigma^2}4 |\nabla w^k |^2\phi \leq \frac{\sigma^2}{8} |\nabla w^{k-1} |^2\phi+C | w^k |^2\phi+(\Delta_k^2+\dt^2)\phi.
\end{align*}
{\color{blue}For $\beta\ge\max\{1,2C\}$, absorption and multiplication
by two yield}
\beq\label{2.10}
\begin{aligned}
&{\beta} | w^k |^2\phi - \frac{1}{2} \Sigma_0 : \nabla^2 ( |w^k|^2 \phi ) +\frac{\sigma^2}2 |\nabla w^k |^2\phi\leq \frac{\sigma^2}{4} |\nabla w^{k-1} |^2\phi+2(\Delta_k^2+\dt^2)\phi.
\end{aligned}
\eeq

{\color{blue}
We next sum the iteration bounds. Define $\mathcal L:=\beta-\tfrac12\Sigma_0:D^2$ and, for $k\ge1$,
\[
S^k:=\phi\sum_{i=1}^k2^{i-k}|w^i|^2,
\qquad
\widetilde\Delta_k:=2\sum_{i=1}^k2^{i-k}\Delta_i^2.
\]
Multiply \eqref{2.10} at iteration $i$ by $2^{i-k}$ and sum over
$1\le i\le k$. The intermediate gradient terms cancel, since
\[
\frac{\sigma^2}{2}\sum_{i=1}^k2^{i-k}|\nabla w^i|^2
-\frac{\sigma^2}{4}\sum_{i=1}^k2^{i-k}|\nabla w^{i-1}|^2
=\frac{\sigma^2}{2}|\nabla w^k|^2
-\frac{\sigma^2}{2^{k+1}}|\nabla w^0|^2.
\]
Using $\sum_{i=1}^k2^{i-k}\le2$, we obtain
\[
\mathcal L S^k+\frac{\sigma^2}{2}|\nabla w^k|^2\phi
\le\frac{\sigma^2}{2^{k+1}}|\nabla w^0|^2\phi
+\widetilde\Delta_k\phi+4\dt^2\phi=:F_k.
\]
The residual assumption and the triangle inequality in $L^d$ give,
including when $d=1$,
\[
\|\widetilde\Delta_k\phi\|_{L^d(B_{2R})}
\le2\sum_{i=1}^k2^{i-k}\|\Delta_i^2\|_{L^d(B_{2R})}
\le4\gamma^2.
\]
Since $\|\nabla w^0\|_\infty\le C$ and
$\|\phi\|_{L^d(B_{2R})}\le C_dR$,
\begin{equation}\label{Fbd}
\|F_k\|_{L^d(B_{2R})}
\le C\bigl(R2^{-k}+\gamma^2+R\dt^2\bigr).
\end{equation}

Apply the ABP estimate \cite[Section~9.1]{gilbarg1977elliptic} to
$\tfrac12\Sigma_0:D^2S^k-\beta S^k\ge-F_k$ on $B_{2R}$:
\[
\sup_{B_{2R}}S^k
\le\sup_{\partial B_{2R}}S^k+CR\|F_k\|_{L^d(B_{2R})}.
\]
The value bound $\|w^i\|_\infty\le2r_{\max}/\beta$, the bound
on the geometric sum, and the boundary estimate for $\phi$ imply
\[
\sup_{\partial B_{2R}}S^k\le Ce^{-2R}.
\]
On $B_R$, $S^k\ge|w^k|^2$. Hence
\[
\sup_{B_R}|w^k|^2
\le C\bigl(e^{-2R}+R^2 2^{-k}+R\gamma^2+R^2\dt^2\bigr)
\tcrev{\le CR^2\bigl(2^{-k}+\gamma^2+\dt^2+e^{-2R}\bigr)}.
\]
\tcrev{The last inequality uses $R\ge1$, with $C$ independent of $R$.
This proves the claim.}
}
\end{proof}
}

\subsection{From Q-evaluation error to a local residual budget}
\label{app:pi-q-to-residual}

{\color{blue}
We bound the local score residual by the Q-evaluation error under
$\mu_{\mathcal K}$. We first establish uniform score-error bounds,
then apply action coverage and interpolation on $B_{2R}$.
Vector norms are Euclidean and matrix norms are operator norms.

\subsubsection{Score-error envelopes}

\begin{lemma}[Local score-error envelopes]
\label{rem:envelope}
Assume \hyperref[ass:phibe-dynamics]{A1}--\tcb{\hyperref[ass:sampling]{A3}},
fix $\beta\ge\beta_{\rm stab}$, and take
$0<\dt\le\min\{1,\Delta t_{\rm reg}\}$.
Work on the event of Proposition~\ref{prop:sw-uniform-matrix} with
\[
r_{A,\mathbb S}^{\mathrm{SE}}(n,\delta;\beta,\dt)
\le\tfrac12\alpha_\Phi(\beta,\dt).
\]
There exist deterministic constants $L_{{\rm imp},R}$ and
$M_{{\rm imp},R}$ such that, for every policy $\pi_j$ on the
algorithmic path, the score error
\[
f_j(s,a):=\beta\bigl(Q_{\dt}^{\pi_j}(s,a)
                 -\widehat Q_{\dt,\Phi}^{\pi_j}(s,a)\bigr)
\]
satisfies, on $B_{2R}\times\mathcal A$,
\[
|f_j(s,a)-f_j(s,a')|\le L_{{\rm imp},R}\|a-a'\|_2,
\qquad |f_j(s,a)|\le M_{{\rm imp},R}.
\]
The constants are independent of the realized batch and iteration index,
but may depend on $\beta,n,\delta,\dt$. Explicit choices are given below.
\end{lemma}

\begin{proof}
By Lemma~\ref{lem:pi-fixed-policy-regularity}, the gradient and
Hessian bounds $G_R(\beta),H_R(\beta)$ hold uniformly along the PI path.

Let $B_{b,R},B_{\Sigma,R}$ be deterministic uniform bounds on
$|b_{\dt}|$ and
$\|\Sigma_{\dt}\|_{\mathrm{op}}$ on $B_{2R}\times\mathcal A$,
respectively. Define
\[
C_\theta(n,\delta;\beta,\dt)
:=\frac{2}{\alpha_\Phi(\beta,\dt)}
\bigl(\|c\|_2+r_c^{\mathrm{SE}}(n,\delta)\bigr),
\]
\begin{align*}
L_{{\rm imp},R}
&:=L_{r,R}+L_{b,R}G_R(\beta)
  +\tfrac d2 L_{\Sigma,R}H_R(\beta)
  +\beta L_{\Phi,R}C_\theta,\\
M_{{\rm imp},R}
&:=r_{\max}+B_{b,R}G_R(\beta)
  +\tfrac d2 B_{\Sigma,R}H_R(\beta)
  +\beta B_\Phi C_\theta.
\end{align*}

On the stated event, the matrix perturbation bound and the definition
of $\alpha_\Phi$ imply
\[
\sup_u\|(\widehat A_{\dt}^{\pi_u})^{-1}\|_{\mathrm{op}}
\le\frac{2}{\alpha_\Phi},\qquad
\|\widehat c\|_2\le\|c\|_2+r_c^{\mathrm{SE}}(n,\delta).
\]
Hence $\|\widehat\theta_j\|_2\le C_\theta$ uniformly along the PI path.

Use the normalized identity
\[
\beta Q_{\dt}^{\pi_j}(s,a)
=r(s,a)+b_{\dt}(s,a)\cdot\nabla V_{\dt}^{\pi_j}(s)
+\tfrac12\Sigma_{\dt}(s,a):D^2V_{\dt}^{\pi_j}(s).
\]
Using $|A:H|\le d\|A\|_{\mathrm{op}}\|H\|_{\mathrm{op}}$,
the coefficient bounds and action-Lipschitz estimates in
Lemma~\ref{lem:standing-regularity}, specifically
\eqref{eq:derived-moment-bounds} and
\eqref{eq:derived-coefficient-action-lipschitz}, give the first three terms of
$M_{{\rm imp},R}$ and $L_{{\rm imp},R}$. A2(a) and
Lemma~\ref{lem:feature-action-regularity} give
\[
|\beta\Phi(s,a)^\top\widehat\theta_j|\le\beta B_\Phi C_\theta,
\qquad
\beta|(\Phi(s,a)-\Phi(s,a'))^\top\widehat\theta_j|
\le\beta L_{\Phi,R}C_\theta\|a-a'\|_2.
\]
Adding the two contributions proves the envelope bounds.
\end{proof}

\subsubsection{Action-max norm transfer}
\label{app:norm-transfer}

By \tcb{A3(b)}, a ball of radius $h$ around a maximizing action has conditional
behavior mass at least $c_h$. The error at that action is therefore
controlled by an average-error term and a Lipschitz correction of size
$hL$. Minimizing over $h$ gives the function $\omega_{{\rm imp},R}$ below.

Write $\|g\|_{L^2(\bar\mu_{\mathcal K};B_{2R})}^2
:=\int_{B_{2R}}|g(s)|^2\bar\mu_{\mathcal K}(s)\,ds$.

\begin{lemma}[Local action-max norm transfer]
\label{lem:action-improvement}
Assume the action-coverage condition in \tcb{\hyperref[ass:sampling]{A3}(b)}.
Let $f:\mathcal K\times\mathcal A\to\mathbb R$ be measurable, and
continuous in the action on $B_{2R}\times\mathcal A$, with
\[
|f(s,a)-f(s,a')|\le L\|a-a'\|_2,
\qquad s\in B_{2R},\quad a,a'\in\mathcal A,
\]
for some $L\ge0$. Set $g(s):=\max_a|f(s,a)|$. Then
\[
\|g\|_{L^2(\bar\mu_{\mathcal K};B_{2R})}
\le\omega_{{\rm imp},R}(\|f\|_{L^2(\mu_{\mathcal K})};L),
\]
where
\[
\omega_{{\rm imp},R}(x;L)
:=\inf_{0<h\le h_A}
\left\{c_h^{-1/2}x+hL\bar\mu_{\mathcal K}(B_{2R})^{1/2}\right\}.
\]
For every fixed finite $L$, $\omega_{{\rm imp},R}(x;L)\to0$
as $x\downarrow0$.
\end{lemma}

\begin{proof}
Compactness of $\mathcal A$ and continuity in the action imply that
$g$ is the supremum over a countable dense action set, hence is
measurable. Fix $0<h\le h_A$. For each $s\in B_{2R}$, choose a
maximizer $a_s$ of $|f(s,\cdot)|$. If $\|a-a_s\|_2<h$, then
$|f(s,a)|\ge(g(s)-Lh)_+$. By action coverage,
\[
\int_{\mathcal A}|f(s,a)|^2\mu_{\mathcal K}(s,a)\,da
\ge c_h(g(s)-Lh)_+^2\bar\mu_{\mathcal K}(s).
\]
Integration over $B_{2R}$ gives
\[
\|(g-Lh)_+\|_{L^2(\bar\mu_{\mathcal K};B_{2R})}
\le c_h^{-1/2}\|f\|_{L^2(\mu_{\mathcal K})}.
\]
Use $g\le(g-Lh)_++Lh$ and take the infimum over $h$.
For continuity at zero, first fix $h$ so that the Lipschitz term is
arbitrarily small, then let $x\downarrow0$ with that $h$ fixed.
\end{proof}

\begin{corollary}[Finite-action norm transfer]
\label{cor:action-improvement-finite}
For a finite action set, replace action integration by counting measure.
If $\mu_{\mathcal K}(s,a)=\bar\mu_{\mathcal K}(s)\nu(a\mid s)$
with $\nu(a\mid s)\ge p_{\min}>0$ on $B_{2R}$, then
\[
\|\max_a|f(\cdot,a)|\|_{L^2(\bar\mu_{\mathcal K};B_{2R})}
\le p_{\min}^{-1/2}\|f\|_{L^2(\mu_{\mathcal K})}.
\]
No action-Lipschitz bound is needed for this conversion.
\end{corollary}
\begin{proof}
Pointwise, $\sum_a|f(s,a)|^2\nu(a\mid s)
\ge p_{\min}\max_a|f(s,a)|^2$. Integrate against
$\bar\mu_{\mathcal K}$ on $B_{2R}$ and take square roots.
\end{proof}

\begin{corollary}[Polynomial action coverage]
\label{cor:action-improvement-poly}
Suppose $c_h\ge c_{\rm ac}h^{q_{\rm ac}}$ for $0<h\le h_A$,
where $c_{\rm ac},q_{\rm ac}>0$. For $L>0$, let
$B_R(L):=L\bar\mu_{\mathcal K}(B_{2R})^{1/2}$ and define
\[
\bar\epsilon_{{\rm imp},R}(L)
:=\frac{2}{q_{\rm ac}}c_{\rm ac}^{1/2}B_R(L)
h_A^{(q_{\rm ac}+2)/2}.
\]
For $0\le x\le\bar\epsilon_{{\rm imp},R}(L)$,
\[
\omega_{{\rm imp},R}(x;L)
\le C(q_{\rm ac})c_{\rm ac}^{-1/(q_{\rm ac}+2)}
B_R(L)^{q_{\rm ac}/(q_{\rm ac}+2)}x^{2/(q_{\rm ac}+2)}.
\]
\end{corollary}
\begin{proof}
Minimize $c_{\rm ac}^{-1/2}xh^{-q_{\rm ac}/2}+B_R(L)h$.
For $x>0$ its unconstrained minimizer is
\[
h_*=\left(\frac{q_{\rm ac}c_{\rm ac}^{-1/2}x}
{2B_R(L)}\right)^{2/(q_{\rm ac}+2)}.
\]
The stated threshold gives $h_*\le h_A$; substitution proves
the bound. For $x=0$, take $h\downarrow0$.
\end{proof}

\begin{remark}
If $L=0$, taking $h=h_A$ gives
$\omega_{{\rm imp},R}(x;0)\le c_{h_A}^{-1/2}x$.
If $c_h\ge c_{\rm ac}>0$ for every $h$, taking $h\downarrow0$
gives $\omega_{{\rm imp},R}(x;L)\le c_{\rm ac}^{-1/2}x$.
\end{remark}

\begin{remark}[Why coverage is applied to the full error]
\label{rem:why-not-c3}
The Galerkin error $Q_{\dt}^{\pi_j}-Q_{\dt,\Phi}^{\pi_j}$ need not
belong to $\operatorname{span}(\Phi)$. A norm equivalence on that span
therefore does not control the full evaluation error. We use
Lemma~\ref{lem:action-improvement} instead.
\end{remark}

\subsubsection{Residual budget along the PI path}
\label{app:pi-residual-budget}

\begin{lemma}[Q-evaluation error to a local residual budget]
\label{lem:pe-to-budget}
Under the assumptions and stability event of Lemma~\ref{rem:envelope},
let $\{\pi_j\}_{j=0}^N$ be the algorithmic path and suppose
\[
\sup_{0\le j<N}
\|Q_{\dt}^{\pi_j}-\widehat Q_{\dt,\Phi}^{\pi_j}\|_{L^2(\mu_{\mathcal K})}
\le\epsilon_Q.
\]
For $1\le k\le N$, the residual in
\eqref{eq:abstract-approx-error} equals
\[
\Delta_k(s)=\max_a|f_{k-1}(s,a)|
=\beta\max_a|Q_{\dt}^{\pi_{k-1}}(s,a)
                  -\widehat Q_{\dt,\Phi}^{\pi_{k-1}}(s,a)|.
\]
It satisfies
\begin{align*}
\|\Delta_k\|_{L^2(\bar\mu_{\mathcal K};B_{2R})}
&\le\omega_{{\rm imp},R}(\beta\epsilon_Q;L_{{\rm imp},R}),\\
\|\Delta_k\|_{L^{2d}(B_{2R})}
&\le\Gamma(\epsilon_Q),
\end{align*}
where
\begin{equation}\label{eq:gamma-R-map}
\begin{aligned}
\Gamma(x)
&:=M_{{\rm imp},R}^{1-1/d}
 m_{\mathcal K,R}^{-1/(2d)}
 \left[\inf_{0<h\le h_A}
 \left\{c_h^{-1/2}\beta x
       +hL_{{\rm imp},R}\bar\mu_{\mathcal K}(B_{2R})^{1/2}\right\}
 \right]^{1/d}\\
&=M_{{\rm imp},R}^{1-1/d}
 m_{\mathcal K,R}^{-1/(2d)}
 \omega_{{\rm imp},R}(\beta x;L_{{\rm imp},R})^{1/d}.
\end{aligned}
\end{equation}
Here $c_h,h_A,m_{\mathcal K,R}$ are the coverage constants in A3(b).
The constants $L_{{\rm imp},R}$ and $M_{{\rm imp},R}$ bound the
action-Lipschitz constant and sup norm of the score error
$\beta(Q_{\dt}^{\pi_j}-\widehat Q_{\dt,\Phi}^{\pi_j})$ on
$B_{2R}\times\mathcal A$, respectively, as constructed in
Lemma~\ref{rem:envelope}.
We suppress the dependence of $\Gamma,L_{{\rm imp},R},M_{{\rm imp},R}$
on $(n,\delta;\beta,\dt)$. These quantities are independent of the
realized batch and iteration index. The map $\Gamma$ is nondecreasing
and tends to zero at zero when these parameters are fixed.
For $d=1$ the factor $M_{{\rm imp},R}^{1-1/d}$ is interpreted as one.
If $M_{{\rm imp},R}=0$, all residuals vanish on $B_{2R}$.
\end{lemma}

\begin{proof}
The state-only term subtracted from the exact and fitted scores
cancels, giving the displayed identity for $\Delta_k$.
Apply Lemma~\ref{lem:action-improvement} to $f_{k-1}$, whose
$L^2(\mu_{\mathcal K})$ norm is at most $\beta\epsilon_Q$ and
whose action-Lipschitz constant is at most $L_{{\rm imp},R}$.
This proves the first bound. \tcb{A3(b)}'s state-density lower bound gives
\[
\int_{B_{2R}}\Delta_k^2\,ds
\le m_{\mathcal K,R}^{-1}
\omega_{{\rm imp},R}(\beta\epsilon_Q;L_{{\rm imp},R})^2.
\]
For $d>1$, use $0\le\Delta_k\le M_{{\rm imp},R}$ to obtain
\[
\int_{B_{2R}}\Delta_k^{2d}\,ds
\le M_{{\rm imp},R}^{2d-2}\int_{B_{2R}}\Delta_k^2\,ds.
\]
Taking the $2d$-th root proves the claim; for $d=1$, the $L^2$ bound
suffices. Since $j=k-1<N$, the same budget holds for every $1\le k\le N$.
\end{proof}

}

\subsection{Assembly of the main-text PI propagation result}
\label{app:pi-assembly}
\label{app:proof-main-pi}

{\color{blue}
We combine Lemma~\ref{lem:pe-to-budget} with
Theorem~\ref{thm:abstract-pi} to prove the PI propagation bound.

\begin{proof}[Proof of Theorem~\ref{thm:q-error-pi-propagation}]
Choose $\Delta t_{\rm PI}\le\min\{1,\Delta t_0\}$, so that
$0<\dt\le\min\{\Delta t_{\rm PI},\Delta t_{\rm reg}\}$
lies in the step-size range of Lemmas~\ref{rem:envelope} and
\ref{lem:standing-regularity}, and take
$\beta\ge\max\{C_0,\beta_{\rm stab}\}$.
Fix $R\ge1$ and work on its empirical stability event.
By Lemma~\ref{lem:pe-to-budget},
\[
\sup_{0\le j<N}
\|Q_{\dt}^{\pi_j}-\widehat Q_{\dt,\Phi}^{\pi_j}\|_{L^2(\mu_{\mathcal K})}
\le\epsilon_Q
\quad\Longrightarrow\quad
\|\Delta_k\|_{L^{2d}(B_{2R})}\le\Gamma(\epsilon_Q),
\qquad 1\le k\le N.
\]
For the surrogate coefficients
$(\widetilde r,\widetilde b,\widetilde\Sigma)
=(r,b_{\dt},\Sigma_{\dt})$, A1 and
Lemma~\ref{lem:standing-regularity} give
\[
\Sigma_{\dt}=\Sigma_0+E_{\dt},\qquad
\|E_{\dt}\|_\infty\le C_e\dt,\qquad
\Sigma_{\dt}\succeq\tfrac12\lambda_0 I
\]
for sufficiently small $\dt$. Lemma~\ref{lem:composed-regularity} supplies the closed-loop
regularity along the path in $\Pi_\Phi$.
\tcrev{Taking $\gamma=\Gamma(\epsilon_Q)$ and $k=N$ in the estimate at the
end of the proof of Theorem~\ref{thm:abstract-pi}, for its stated
discount and step-size range, yields}
\[
\sup_{s\in B_R}|V_{\dt}^{\pi_N}(s)-V_{\dt}^*(s)|^2
\tcrev{\le C\left(R^2 2^{-N}+R\Gamma(\epsilon_Q)^2+R^2\dt^2+e^{-2R}\right)},
\]
\tcrev{where $C$ is independent of $R$.}
Since $\varrho$ is a probability density supported in $B_R$,
\[
\|V_{\dt}^{\pi_N}-V_{\dt}^*\|_\varrho^2
\le
\sup_{s\in B_R}|V_{\dt}^{\pi_N}(s)-V_{\dt}^*(s)|^2.
\]
\tcrev{Taking square roots and using $R\ge1$ gives}
\[
\tcrev{\begin{aligned}
\|V_{\dt}^{\pi_N}-V_{\dt}^*\|_\varrho
&\le C_{\rm prop}
\left[R2^{-N/2}+\sqrt R\,\Gamma(\epsilon_Q)+R\dt+e^{-R}\right]\\
&\le C_{\rm prop}R
\left[2^{-N/2}+\Gamma(\epsilon_Q)+\dt+e^{-R}\right].
\end{aligned}}
\]
\tcrev{The constant $C_{\rm prop}$ inherits only the global coefficient and
regularity dependence of the contraction estimate and is independent
of $R$. The sample-size, coverage, and envelope dependence remains in
$\Gamma(\epsilon_Q)$, which may also depend on $R$.}
This proves the stated propagation bound.
\end{proof}

\begin{remark}[Action-independent working diffusion]
If $\Sigma_{\dt}(s,a)\equiv\Sigma_0(s)$, then $E_{\dt}=0$.
The diffusion-perturbation term in the contraction proof vanishes, and
the same argument gives
\[
\|V_{\dt}^{\pi_N}-V_{\dt}^*\|_\varrho
\tcrev{\le C_{\rm prop}R
\left[2^{-N/2}+\Gamma(\epsilon_Q)+e^{-R}\right]}.
\]
\end{remark}
}

\end{document}